\documentclass[sigconf]{acmart}

\usepackage{booktabs} 

\usepackage{caption}

\DeclareCaptionLabelFormat{lc}{{#1}~#2}
 \usepackage{epsfig,endnotes}

\usepackage{floatpag,enumitem}
\usepackage{hyphenat}
\setlist{leftmargin=10pt}
 \usepackage{relsize}

\newcommand{\de}{\stackrel{\text{def}}{=}}
\newcommand{\con}{\boldsymbol{\mid}}
\usepackage{multirow}
\usepackage{setspace}
\usepackage{mathrsfs}
\usepackage{bbm}
\usepackage{pifont}
\usepackage{amsfonts}

\usepackage{amsmath, amsthm}

\usepackage{tabularx}
\usepackage{listings}
\usepackage{graphicx}
\usepackage{amssymb}
\usepackage{array}
\usepackage{fancyhdr}
\usepackage{cases}

 \usepackage{supertabular}
\usepackage{url}
\usepackage{hyperref}
\usepackage{fancyhdr}
\usepackage{mathrsfs}
\newcommand{\pr}{\mathbb{P}} 

\usepackage{psfrag}
\usepackage{bbding}

\newcommand{\bcup} {\hspace{2pt} \mathlarger{\cup}
\hspace{2pt}}\newcommand{\h}{it holds that }

\newcommand{\f}{it follows that }

\usepackage{float}

\usepackage{amsbsy}

\makeatletter
\providecommand{\leftsquigarrow}{%
  \mathrel{\mathpalette\reflect@squig\relax}%
}
\newcommand{\reflect@squig}[2]{%
  \reflectbox{$\m@th#1\rightsquigarrow$}%
}
\makeatother

\usepackage{algorithm}
 \usepackage{algorithmic}

\makeatletter
\newcommand{\newalgname}[1]{%
  \renewcommand{\ALG@name}{#1}%
}

\newcommand{\bp}[1]{{\mathbb{P}}\left[{#1}\right]}

\newcommand{\fsquare}{\vrule height6pt width7pt depth1pt}   
\newcommand{\pfe}{\hfill\fsquare}             

\def\centerhack#1{\hbox to 0pt{\hss\footnotesize #1\hss}}
\def\centerhackn#1{\hbox to 0pt{\hss #1\hss}}
\def\dchack#1{\vbox to 0pt{\vss{\hbox to 0pt{\hss#1\hss}}\vss}}

\newtheorem{lem}{Lemma}
\newtheorem{thm}{Theorem}

\newtheorem*{proposition1.1}{Proposition 1.1}
\newtheorem*{proposition1.2}{Proposition 1.2}
\newtheorem*{proposition1.3}{Proposition 1.3}
\newtheorem*{proposition2.1}{Proposition 2.1}
\newtheorem*{proposition2.2}{Proposition 2.2}
\usepackage{subfigure}

\setcopyright{acmcopyright}
\acmDOI{10.1145/3139550.3139556}
\acmISBN{ISBN: 978-1-4503-5175-1/17/10}
\acmConference[WPES'17, October 30, 2017]{ACM Workshop on Privacy in the Electronic Society (WPES), held in conjunction with ACM Conference on Computer and Communications Security (CCS)}{October 30, 2017}{Dallas, Texas, USA}
\acmYear{2017}
\copyrightyear{2017} \acmPrice{15.00}

\begin{document}
\title{Adaptive Statistical Learning with Bayesian Differential Privacy}

\author{Jun Zhao}
\affiliation{%
  \institution{Carnegie Mellon University and Nanyang Technological University$^*$\thanks{$^*$The author Jun Zhao obtained his PhD from Carnegie Mellon University, \mbox{Pittsburgh,}  PA 15213, USA, where he was with the Cybersecurity Lab (CyLab). He was a postdoctoral scholar with Arizona State University, Tempe, AZ 85281, USA. He is now a research fellow at Nanyang Technological University in Singapore.
 Email: \vspace{5pt} \texttt{junzhao@alumni.cmu.edu}}}
}
\email{junzhao@alumni.cmu.edu}

\renewcommand{\shortauthors}{J. Zhao}

\begin{abstract}

In statistical learning, a dataset is often partitioned   into two parts: the training set and the
holdout (i.e., testing) set. For instance, the training set is used to learn a predictor, and then the holdout set is used for
estimating the accuracy of the predictor on the true distribution. However, often in practice,
 the holdout dataset is
reused  and the estimates tested on the holdout dataset are
chosen \textit{adaptively} based on the results of prior estimates, leading to that the predictor may become dependent
of the holdout set. Hence, \textit{overfitting} may occur, and the learned models may \textit{not generalize} well to the unseen datasets. Prior studies have established connections between
the stability of a learning algorithm and its ability to generalize, but the traditional generalization is not
robust to \textit{adaptive} composition. Recently, Dwork~\emph{et~al}. in NIPS, STOC, and Science 2015 show that the holdout dataset from \textit{i.i.d. data samples} can be \textit{reused} in \textit{adaptive statistical learning}, if the estimates are
perturbed and coordinated using techniques developed for \textit{differential privacy}, which is a widely used notion to quantify privacy. Yet, the results of Dwork~\emph{et~al}.  are applicable to only the case of i.i.d. samples. In contrast, correlations between data samples exist because of various behavioral, social, and genetic relationships between users. Our results in adaptive statistical learning generalize the results of Dwork~\emph{et~al}. for \textit{i.i.d. data samples} to \textit{arbitrarily correlated data}. Specifically, we show that the holdout dataset from correlated samples can be \textit{reused} in adaptive statistical learning, if the estimates are
perturbed and coordinated using techniques developed for \textit{Bayesian differential privacy}, which is a privacy notion recently introduced by Yang~\textit{et~al.} in SIGMOD 2015 to broaden  the application scenarios of differential privacy when data records are correlated.

\end{abstract}

%
%



\maketitle

\pagestyle{plain}

\thispagestyle{plain}

\section{Introduction }

In many statistical learning algorithms, a common practice is to partition a dataset  into two parts: the training set and the
holdout (i.e., testing) set. For instance, after the training set is used to learn a predictor,  the holdout set is used for
estimating the accuracy of the predictor on the true distribution. However, in many practical applications, since (i)
 the holdout dataset is
reused, and (ii) the estimates tested on the holdout dataset are
chosen \textit{adaptively} based on the results of prior estimates, we observe that the predictor may become dependent
of the holdout set. This leads to the result that \textit{overfitting} may occur, and the learned models may \textit{not generalize} well to the unseen datasets. Several papers \cite{bousquet2002stability,mukherjee2006learning,poggio2004general,shalev2010learnability} in the literature have established connections between
the stability of a learning algorithm and its ability to generalize, but the traditional generalization is not
robust to \textit{adaptive} composition \cite{bassily2016typicality,cummings2016adaptive,dwork2015preserving}. To remedy this issue, Dwork \emph{et al}. \cite{dwork2015generalizationarXiv,dwork2015preserving,dwork2015Sciencereusable} recently show that the holdout dataset from i.i.d. data samples can be \textit{reused} in \textit{adaptive} statistical learning, if the estimates are
perturbed and coordinated using techniques developed for \textit{differential privacy}, which has emerged as the standard notion to quantify privacy and will be elaborated next.

\textit{Differential privacy} by Dwork \textit{et al.} \cite{Dwork2006,dwork2006calibrating} is a   privacy notion that has been successfully
applied to a range of statistical learning tasks, since it offers a rigorous foundation for defining privacy. Differential privacy has received considerable interest in the literature \cite{blocki2016differentially,xiao2015protecting,zhang2016privtree,loucost,wang2017privsuper,shokri2015privacy}. The   Chrome browser by Google has used a differentially private tool called
RAPPOR \cite{erlingsson2014rappor} to collect information about clients. Starting from iOS 10, Apple~\cite{apple} has incorporated differential privacy into its mobile operating system.
 A randomized mechanism $Y$ satisfies $\epsilon$-differential privacy
if for all neighboring databases $x$ and $x'$ that differ in one record, and for any subset $\mathcal{Y}$ of the output range of the mechanism $Y$, it holds that
\mbox{$\pr[Y(x) \in \mathcal{Y}] \leq e^{\epsilon} \pr[Y(x') \in \mathcal{Y}],$} where $\pr[\cdot]$ denotes the probability and $e$ is a mathematical constant that is the base of the natural logarithm.  Intuitively, under differential privacy, an adversary
given access to the output does not have much confidence to determine
whether the output was sampled from the probability distribution
generated by the randomized algorithm when the database is $x$ or when the database is~$x'$ that is different from $x$ by one record.

Dwork \emph{et al}. \cite{dwork2015generalizationarXiv,dwork2015preserving,dwork2015Sciencereusable} show that the holdout dataset from \textit{i.i.d. data samples} can be \textit{reused} in adaptive statistical learning, if the estimates are
perturbed and coordinated using algorithms satisfying differential privacy. These efforts by Dwork \emph{et al}. \cite{dwork2015generalizationarXiv,dwork2015preserving,dwork2015Sciencereusable} have attracted significant attention to adaptive statistical learning \cite{bassily2016typicality,blum2015ladder,cummings2016adaptive,rogers2016max,russo2016controlling}. However, existing studies for adaptive statistical learning including those of Dwork \emph{et al}. \cite{dwork2015generalizationarXiv,dwork2015preserving,dwork2015Sciencereusable} are applicable to \textit{only the case of i.i.d. samples}. In contrast, correlations between data samples exist because of various behavioral, social, and genetic relationships between users \cite{BDPIT}. In location privacy, a user's locations across time exhibit temporal correlations \cite{olteanu2016quantifying,xiao2015protecting,wang2016privacy},  and locations of friends tend to have social correlations since they are likely to visit the same place \cite{Changchang2016,backstrom2010find}. In genome privacy, DNA information is passed from parents to children based on Mendelian inheritance  so family members' genotypes are correlated, where a genotype is the set of genes in DNA  responsible for a particular trait \cite{humbert2013addressing}.
In short, real-world data samples may contain a rich set of correlations.

Our results in adaptive statistical learning generalize recent work of Dwork \emph{et al}. \cite{dwork2015generalizationarXiv,dwork2015preserving,dwork2015Sciencereusable} for \textit{i.i.d. data samples} to \textit{arbitrarily correlated data}. In other words, we tackle adaptive statistical learning with \textit{correlated data samples} while Dwork \emph{et al}. \cite{dwork2015generalizationarXiv,dwork2015preserving,dwork2015Sciencereusable} consider only \textit{i.i.d. samples}. Specifically, we show that the holdout dataset from correlated samples can be \textit{reused} in adaptive statistical learning, if the estimates are
perturbed and coordinated using techniques developed for \textit{Bayesian differential privacy}, a privacy notion recently introduced by Yang \textit{et al.}  \cite{yang2015bayesian} to address correlations in a database. It has   been observed by Kifer and Machanavajjhala \cite{kifer2011no} (see also \cite{chen2014correlated,he2014blowfish,kifer2012rigorous,Changchang2016,wang2016privacy,zhu2015correlated,BDPIT}) that differential privacy may not work as expected when the data tuples are correlated. The underlying reason according to Zhao~\emph{et~al.}~\cite{BDPIT} is that differential privacy's guarantee masks only the presence of those records received \textit{from} each user, but could not mask  statistical trends that may reveal information \textit{about} each user. Bayesian differential privacy  \cite{yang2015bayesian} broadens the application scenarios of differential privacy when data records are correlated. For clarity, we defer the detailed definition of Bayesian differential privacy to Section~\ref{preliminaries:BDP}.

The rest of the paper is organized as follows. We discuss some preliminaries in Section
\ref{sec:preliminaries}, before presenting our results of adaptive statistical learning in Section
\ref{sec:main:res}. Section~\ref{sec:Experiments} provides experiments  to support our results. We elaborate the proofs in Section
\ref{sec:Proofs}.  Section \ref{related} surveys related work, and
Section \ref{sec:Conclusion} concludes the paper.

\section{Preliminaries} \label{sec:preliminaries}

\subsection{Adaptive Statistical Learning} \label{preliminaries:ASL}

 Dwork \emph{et al}. \cite{dwork2015generalizationarXiv} show that differential privacy techniques can be leveraged in adaptive statistical learning for i.i.d. data samples. To this end, they   introduce the notion of \textit{approximate max-information},  establish a connection of   differential privacy to approximate max-information, and utilize this connection to present an algorithm for adaptive statistical learning.

For two random variables $X$ and $Y$, let $X \times Y$ be the random variable obtained by drawing
$X$ and $Y$ \emph{independently} from their probability distributions, and let $\textrm{domain}(X,Y)$ be the domain of~$(X,Y)$. \mbox{Dwork~\emph{et al}.~\cite{dwork2015generalizationarXiv}}  define the notion of  \emph{$\beta$-approximate max-information} as follows:
\begin{align}
I_{\infty}^{\beta}(X;Y)\de \log \max\limits_{\begin{subarray}{l}\mathcal{O}\subseteq \textrm{domain}(X,Y):\\ \bp{(X,Y)\in \mathcal{O}}>\beta\end{subarray}} \frac{\bp{(X,Y)\in \mathcal{O}}-\beta}{\bp{X \times Y\in \mathcal{O}}},\label{defIinf}
\end{align}
  where $\beta>0$ and $\log $ means the binary logarithm. Approximate max-information gives generalization since it upper bounds the probability of ``bad events'' that can occur as a result of the dependence of the learning result $Y(X)$ on the dataset $X$; see ~\cite[Page 10]{dwork2015generalizationarXiv} for more details. Specifically, it is straightforward to obtain from (\ref{defIinf}) that if $$I_{\infty}^{\beta}(X;Y(X)) =k,$$  then $$\bp{(X,Y)\in \mathcal{O}} \leq 2^k \cdot \bp{X \times Y\in \mathcal{O}} + \beta$$ for any $\mathcal{O}\subseteq \textrm{domain}(X,Y)$.

Statistical learning considered in this paper is as follows: For an unknown distribution $\mathcal{D}$ over a discrete
universe $\mathcal{X}$ of possible data points, a statistical query
$Q$ asks for the expected value of some function $f: \mathcal{X} \rightarrow [0,1]$ on random draws from $\mathcal{D}$. The goal is to ensure that the estimate obtained from data is close to the true result on the unknown distribution. We consider queries $Q_1,Q_2,\ldots$ that are chosen adaptively based on the results of prior estimates.

\subsection{Differential Privacy} \label{preliminaries:DP}

A randomized algorithm $Y$ satisfies $\epsilon$-differential privacy (DP)
if for all neighboring databases $x$ and $x'$ that differ in one record, and for any subset $\mathcal{Y}$ of the output range of the mechanism $Y$, it holds that
\begin{align}
\pr[Y(x) \in \mathcal{Y}] \leq e^{\epsilon} \pr[Y(x') \in \mathcal{Y}].  \label{DPdef}
\end{align}
 To ensure
$\epsilon$-differential privacy,
the Laplace mechanism \cite{dwork2006calibrating} and the exponential mechanism \cite{mcsherry2007mechanism} have been proposed in the literature. The details are as follows.
\begin{itemize}
\item To achieve
$\epsilon$-differential privacy for a query $Q$, the Laplace mechanism $\text{Lap}({\Delta_Q}/{\epsilon})$ adds Laplace noise with parameter (i.e., scale) ${\Delta_Q}/{\epsilon}$ independently to each dimension of the query result, where $\Delta_Q$ is the {global sensitivity}  of query $Q$:\\$\Delta_Q = \max_{\textrm{neighboring $x,x'$}} \|Q(x) - Q(x')\|_{1}$, where databases~$x$ and~$x'$ are neighboring if they differ in one record.
\item To ensure
$\epsilon$-differential privacy for a query $Q$, the exponential mechanism $\text{Expo}(\epsilon,u,\mathcal{R})$ for some
utility function $u$ and the output range $\mathcal{R}$ outputs an element $y \in \mathcal{R}$ with
probability proportional to $\exp\big(\frac{\epsilon \cdot u(x,y)}{2 \Delta_u}\big)$, where $\Delta_u$ is  the sensitivity of $u$ with
respect to its database argument:\\$\Delta_u = \max_{y \in \mathcal{R}} \max_{\textrm{neighboring $x,x'$}} |u(x,y) - u(x',y)|$, where data-
bases $x$ and $x'$ are neighboring if they differ in one record.
\end{itemize}

Although differential privacy (DP) has been recognized as a powerful notion, it has   been observed by Kifer and Machanavajjhala \cite{kifer2011no} (see also \cite{chen2014correlated,he2014blowfish,kifer2012rigorous,Changchang2016,wang2016privacy,zhu2015correlated,BDPIT}) that DP may not work as expected when the data tuples are correlated. As noted by Zhao~\emph{et~al.}~\cite{BDPIT}, although DP ensures that a user's participation itself in the computation reveals no further secrets from the user, however, due to tuple correlation, a user's data may impact other users' records and hence has more influence on the query response than what is expected compared with the case where   tuples are independent. When correlations exist, a user's data is not known to the user alone in some degree, and an adversary may combine the query output and the correlation to learn about a
user's data.

To extend differential privacy for correlated data, prior studies have investigated
various privacy metrics \cite{chen2014correlated,he2014blowfish,kifer2012rigorous,Changchang2016,wang2016privacy,zhu2015correlated}. One of the metrics receiving much attention is the notion of Bayesian differential privacy  introduced by Yang \textit{et al.}  \cite{yang2015bayesian} as follows.

\begin{figure*}[!tb]
 \vspace{0mm} \centerline{\includegraphics[width=1\textwidth]{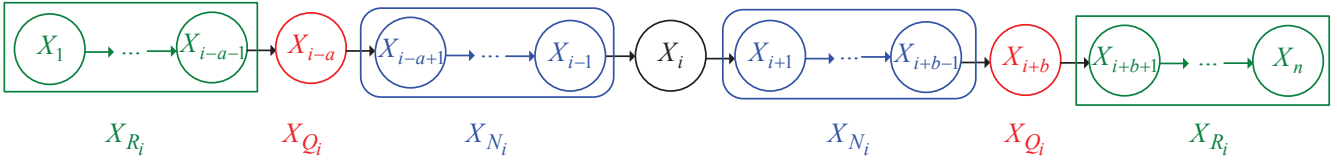}}
\vspace{-5pt}
 \caption{This figure from \cite{BDPIT} is based on \cite{wang2016privacy}. For a Markov chain $X_1 \to X_2 \ldots \to X_n$, moralization simply means making each directed edge undirected, since there are no nodes that have a common child. A \textit{Markov quilt} for node $X_i$ is $X_{Q_i}=\{X_{i-a},X_{i+b}\}$. The corresponding \textit{nearby set} $X_{N_i}$ and \textit{remote set} $X_{R_i}$ are $X_{N_i}=\{X_{i-a+1},\ldots,X_{i-1}\}\bcup \{X_{i+1},\ldots,X_{i+b-1}\}$ and $X_{R_i}=\{X_1,\ldots,X_{i-a-1}\}\bcup \{X_{i+b+1},\ldots,X_n\}$. To check the definition of a Markov quilt, it is straightforward to see that here any path between $X_i$ and a node in $X_{R_i}$ has to include at least one node in $X_{Q_i}$. Both $a$ and $b$ above are positive integers. If $a=1$ and $b=1$, the \textit{Markov quilt} becomes the \textit{Markov blanket} $X_{M_i}=\{X_{i-1},X_{i+1}\}$ with the corresponding $X_{N_i}$  being $\emptyset$. Note that any $X$ term with an index outside of $\{1,2,\ldots,n\}$ is canceled out; e.g., the Markov blanket $X_{M_1}$ of node $X_1$ is actually $\{X_2\}$ rather than $\{X_0,X_2\}$ since there is no node $X_0$.\vspace{0pt}}
\label{fig-Markovblanketquilt}
\end{figure*}

\subsection{Bayesian Differential Privacy} \label{preliminaries:BDP}

In this paper, we will establish a connection of Bayesian differential privacy to approximate max-information and then leverage this connection to use Bayesian differential privacy for adaptive statistical learning. We discuss Bayesian differential privacy below.

 The notion of Bayesian differential privacy (BDP) is introduced by Yang \textit{et al.}  \cite{yang2015bayesian} to extend differential privacy for addressing the case when data records are correlated. Before stating the definition, we first introduce some notation.

The database under  consideration is modeled by a random variable $X=[{X}_1,{X}_2, \ldots, {X}_n]$, where ${X}_j$ for each $j\in  \{1,\ldots,n\}$ is a \emph{tuple}, which is also a \emph{random variable}. Let database $x=[x_1,x_2, \ldots,x_n]$ be an instantiation of $X$, so that each ${x}_j$ denotes an instantiation of ${X}_j$. Let $i$ be the index of the tuple attacked by the adversary.
For notational simplicity, $x_S$ and $X_S$ stand for $[x_j\hspace{-1pt}:j\hspace{-1pt}\in\hspace{-1pt} S]$ and $[X_j\hspace{-1pt}:j\hspace{-1pt}\in\hspace{-1pt}S]$ respectively for any set $S \hspace{-1pt}\subseteq \hspace{-1pt}\{1,2,\ldots,n\}\setminus \{i\}$; i.e., $x_S$ is an instantiation of ${X}_S$.
 An
adversary denoted by $A(i, S) $    knows
the values of all tuples in $S$ (denoted by $x_S$)  and attempts
to attack the value of tuple $i$ (denoted by $x_i$). For a randomized perturbation mechanism $Y$ that maps a database $x$ to a randomized output~$y$, the Bayesian differential privacy leakage (BDPL) of $Y$ with respect to the adversary  $A(i, S) $ is defined by
\begin{align}
\text{BDPL}_A(Y) = \text{sup}_{x_i, x_i', x_S,  \mathcal{Y}} \ln \frac{\pr[y \in \mathcal{Y} \con x_i, x_{S}]}{\pr[y \in \mathcal{Y} \con x_i', x_S]}, \label{BDPL}
\end{align}
where the subscript $A$ is short for $A(i, S) $, and $\ln$ denotes the natural logarithm.
In (\ref{BDPL}), $x_i$ and $x_i'$ iterate through the domain of tuple $X_i$ (i.e., $x_i \in \textrm{domain}(X_i)$, $x_i' \in \textrm{domain}(X_i)$, and $x_i \neq x_i' $), and   $x_{S}$ iterates through the domain of tuple(s) $X_{S}$ (i.e., $x_{S} \in \textrm{domain}(X_{S})$), where $X_S$ and $x_S$ stand for $[X_j:j\in S]$ and $[x_j:j\in S]$ respectively for  $S \subseteq \{1,2,\ldots,n\}\setminus \{i\}$. In (\ref{BDPL}), $\mathcal{Y}$ iterates through all subsets of the output range of the mechanism~$Y$.

The mechanism $Y$ satisfies $\epsilon$-Bayesian differential privacy if
\begin{align}
\text{sup}_A \text{BDPL}_A(Y) \leq \epsilon.  \label{BDPL2}
\end{align}
In (\ref{BDPL2}), $A$ (short for $A(i, S) $) iterates through the set of all adversaries; i.e., $i$ iterates through the index set $ \{1,\ldots,n\}$ and $S$ iterates through all subsets of $ \{1,2,\ldots,n\}\setminus \{i\}$.

Based on (\ref{DPdef})--(\ref{BDPL2}), Yang \textit{et al.}  \cite{yang2015bayesian} show that
when all tuples are independent, (\ref{DPdef}) (i.e., DP guarantee) and (\ref{BDPL2}) (i.e., BDP guarantee) are the same. However, under tuple correlations, (\ref{DPdef}) (i.e., DP guarantee) and (\ref{BDPL2}) (i.e., BDP guarantee) are different due to the following reason \cite{BDPIT}: although DP protects the information received \textit{from} each user, an adversary may combine the correlations and the query response to obtain a large amount of information \textit{about} the user. In other words, under DP, although the user itself is privacy-aware in participating the database, the participations of other users together with the query response nevertheless leak the user's data. The BDP notion ensures that even under tuple correlation, almost no sensitive information about any user can be leaked because of answering the query.

\subsection{Mechanisms to Achieve Bayesian Differential Privacy} \label{preliminaries:BDPmechanisms}

Yang \emph{et al}. \cite{yang2015bayesian} extend the Laplace mechanism \cite{dwork2006calibrating} of differential privacy to achieve Bayesian differential privacy. However, this mechanism of \cite{yang2015bayesian} is only for the sum query on a Gaussian Markov random field (GMRF) with positive correlations and its extension to a discrete domain, so it cannot apply to queries other than the sum query and cannot apply to correlations other than those of positive-correlated GMRF. Recently, my co-authors and I \cite{BDPIT} present mechanisms for arbitrary correlations by connecting Bayesian differential privacy (BDP) to differential privacy (DP). Specifically, we \cite{BDPIT} show that $\epsilon'$-DP implies $\epsilon$-BDP, where $\epsilon'$ depends on $\epsilon$ and the correlations between the data tuples. Note that although \cite{BDPIT} uses the notion of dependent differential privacy, this notion is equivalent to Bayesian differential privacy.

For clarity, we will present \cite{BDPIT}'s results on the relationship between Bayesian differential privacy (BDP) and differential privacy (DP) as Lemmas \ref{thmDPBDP-main-result-generalized}--\ref{lemexactformoffunctionhdpddp} in Section \ref{sec:Useful} later. In order to state these results as well as our main results on adaptive statistical learning in Section \ref{sec:main:res}, we first review some preliminaries given in \cite{BDPIT}.

\textbf{Representing dependency structure of tuples of a database via probability graphical models.} To represent dependency structure of tuples in a database, we \cite{BDPIT} apply the well-known notion called \textit{probability~graphical model}. These models use graphs (i.e., networks) to express the conditional (in)dependencies between random variables. In our
applications, each node stands for a tuple of the database (a tuple is also a random variable), and we have a network to represent the conditional (in)dependencies between tuples of the database.

A Bayesian network (which is a directed acyclic graph) and a Markov network (which is an undirected graph) are two kinds of probability graphical models that have been studied extensively and used in various applications \cite{koller2009probabilistic}. A Bayesian network is a directed acyclic graph which represents a factorization of the joint probability of all random variables.  Specifically, in a Bayesian network of $n$ nodes $\{X_1,X_2,\ldots,X_n\}$, with $\textrm{\texttt{PA}}_i$ denoting the set of parents of node $X_i$  (i.e., each node of $\textrm{\texttt{PA}}_i$ points directly to node $X_i$ via a single directed edge), then the joint probability satisfies $\bp{X_1,X_2,\ldots,X_n}=\prod_{i=1}^n \bp{X_i \con \textrm{\texttt{PA}}_i}$.  A Markov network represents (in)dependencies between random variables via an  undirected graph which has the Markov property such that any two subsets of variables are conditionally independent given a separating subset; i.e., with $A,B,C$ denoting three disjoint sets of nodes, if any path between any node $a \in A$ and any node $b \in B$ has to include at least one node in $C$ (or there is simply no path between $a \in A$ and $b \in B$), then $A$ and $B$ are conditionally independent given $C$.

\textbf{Markov blanket.} \label{defMarkovblanket}  The notion of Markov blanket is standard in probability graphical models \cite{koller2009probabilistic}. For a node $X_i$, its Markov blanket $X_{M_i}$ comprises the tuples that are \emph{directly} correlated with tuple $X_i$ (i.e., given $X_{M_i}$ which contains $X_j$ for $j\in M_i$, $X_i$ is conditionally independent of everything else).  If $X_i$ is independent of all other $n-1$ tuples, then $M_i=\emptyset$.
 In a Bayesian network, a node's Markov blanket consists of its parents, children, and its children's other parents. In a Markov network (also known as a Markov random field), the Markov blanket of a node is its set of neighboring nodes.

\textbf{Markov quilt.} \label{def-Markov-quilt-formal} We \cite{BDPIT} generalize the notion of \textit{Markov blanket} to \emph{Markov quilt}. This definition is adopted from Song~\emph{et~al}.~\cite{wang2016privacy} with slight changes. We consider a network of $n$ nodes $X_1, X_2, \ldots, X_n$. First, we perform moralization if necessary. In other words, for a Bayesian network, we moralize it into a Markov network; 
 for a Markov network, no action is taken. Moralization means making each directed edge undirected and putting an edge between any two nodes that have a common child. Recall that $X_S$ represents the set of nodes with indicies in an index set $S$, where $S \subseteq \{1,2,\ldots,n\}$; i.e. $X_S = \{X_j:j\in S\}$. Let $Q_i$ and $R_i$ be disjoint subsets of \mbox{$\{1,2,\ldots,n\}\setminus \{i\}$.}  We say a set $X_{Q_i}$ of nodes is a \textit{Markov quilt} of node $X_i$ if after moralization (whenever necessary) of the dependency network, for any node $X_j$ in $X_{R_i}$, \textit{either} any path between $X_i$ and $X_j$ has to include at least one node in $X_{Q_i}$, \textit{or} there is simply no path between $X_i$ and $X_j$. We emphasize that this definition is defined on a Markov network or after we have moralized a Bayesian network into a Markov network. The definition implies that $X_{R_i}$ is independent of $X_i$ conditioning on $X_{Q_i}$. Excluding $X_i$, $X_{Q_i}$ and $X_{R_i}$, we define the remaining nodes as $X_{N_i}$; i.e., $X_{Q_i}$, $X_{N_i}$, $X_{R_i}$ together constitute a partition of $\Big\{X_j:j\in\{1,2,\ldots,n\}\setminus \{i\}\Big\}$ so that $Q_i,R_i,N_i$ together partition $\{1,2,\ldots,n\}\setminus \{i\}$. Intuitively, $X_{Q_i}$ separates $X_i$ and $X_{R_i}$, so $X_{R_i}$ is \emph{remote} from $X_i$ while $X_{N_i}$ is \emph{nearby} from $X_i$ (this is why we use the notation $R$ and $N$). We refer to $X_{N_i}$ (resp., $X_{R_i}$) as the \textit{nearby set} (resp., the \textit{remote set}) associated with the \textit{Markov quilt} $X_{Q_i}$. Note that when we define a Markov quilt $X_{Q_i}$ for node $X_i$, we actually have a triple $(X_{Q_i},X_{N_i},X_{R_i})$: a \textit{Markov quilt} $X_{Q_i}$, a \textit{nearby set} $X_{N_i}$, and a \textit{remote set} $X_{R_i}$. Figure \ref{fig-Markovblanketquilt} provides  an illustration of $X_{Q_i}$, $X_{R_i}$, and $X_{N_i}$ on a Markov chain. Clearly, for a node, its Markov blanket is a special Markov quilt. Yet, while a node has only one Markov blanket,
a node may have many different Markov quilts, as presented in Figure \ref{fig-Markovblanketquilt}. \label{defquilt}


\textbf{Max-influence.}
To quantify how
much changing a tuple can impact
other tuples, very recently, Song \emph{et al}. \cite{wang2016privacy} define the \emph{\mbox{max-influence}} of a variable $X_i$ on a
set of variables $X_S$ (i.e., the set of $X_j$ for $j\in S$) for  $S\subseteq  \{1,\ldots,n\} \setminus \{i\}$ as follows:
 \begin{align}
 \mathcal{I}(X_{S}\leftsquigarrow X_i) &  \de \ln \max_{x_{S},x_i,x_i'} \big\{{\bp{x_{S}\con x_i}}/{\bp{x_{S}\con x_i'}}\big\}. \label{defIXMiKXiXKSdefSong}
\end{align}
We \cite{BDPIT} generalize this definition to describe  the \emph{max-influence} of $X_i$ on  a
set of variables $X_S$ \textit{conditioning} on a
set of variables $X_K$ as follows ($S$ and $K$ are disjoint subsets of $ \{1,\ldots,n\} \setminus \{i\}$):
\begin{align}
 \mathcal{I}(X_{S}\leftsquigarrow X_i\con X_K)   &  \de \ln \max_{x_{S},x_i,x_i',x_K}\big\{{\bp{x_{S}\con x_i,x_K}}/{\bp{x_{S}\con x_i',x_K}}\big\}.\label{defIXMiKXiXKSdef}
\end{align}
In (\ref{defIXMiKXiXKSdefSong}) and (\ref{defIXMiKXiXKSdef}), $X_S$ and $x_S$ stand for $[X_j:j\in S]$ and $[x_j:j\in S]$ respectively for  $S\subseteq  \{1,\ldots,n\} \setminus \{i\}$; similarly, $X_K$ and $x_K$ stand for $[X_j:j\in K]$ and $[x_j:j\in K]$ respectively for  $K\subseteq   \{1,\ldots,n\} \setminus \{i\}$. We also have that $S$ and $K$ are disjoint subsets of $ \{1,\ldots,n\} \setminus \{i\}$); $x_i$ and $x_i'$ iterate through the domain of tuple $X_i$ (i.e., $x_i \in \textrm{domain}(X_i)$, $x_i' \in \textrm{domain}(X_i)$, and $x_i \neq x_i' $);  $x_{S}$ iterates through the domain of tuple(s) $X_{S}$ (i.e., $x_{S} \in \textrm{domain}(X_{S})$); and $x_{K}$ iterates through the domain of tuple(s) $X_{K}$ (i.e., $x_{K} \in \textrm{domain}(X_{K})$).

Based on (\ref{defIXMiKXiXKSdef}), $\mathcal{I}(X_{S}\leftsquigarrow X_i\con X_K)$ equals $0$ if and only if $X_{S}$ is independent of $X_i$ conditioning on $X_K$, given the following:
\begin{itemize}
\item On the one hand, if $X_{S}$ is independent of $X_i$ conditioning on $X_K$, then the nominator ${\bp{x_{S}\con x_i,x_K}}$ and the denominator ${\bp{x_{S}\con x_i',x_K}}$ in  (\ref{defIXMiKXiXKSdef}) are the same, so $\mathcal{I}(X_{S}\leftsquigarrow X_i\con X_K)$ becomes $0$.
\item On the other hand, $\mathcal{I}(X_{S}\leftsquigarrow X_i\con X_K)$ equals $0$ only if  ${\bp{x_{S}\con x_i,x_K}}$\\$={\bp{x_{S}\con x_i',x_K}}$ for any $x_i$, any $x_i'$, any $x_{S}$, and any $x_{K}$ (i.e., only if $X_{S}$ is independent of $X_i$ conditioning on $X_K$).
\end{itemize}

In (\ref{defIXMiKXiXKSdef}), we note that $S$ and $K$ are disjoint subsets of $  \{1,\ldots,n\}\setminus\{i\}$.
Also, for generality, if $S=\emptyset$, then $\mathcal{I}(X_{S}\leftsquigarrow X_i\con X_K) = 0$.

\begin{algorithm}
\caption{A generalization-achieving algorithm (based on \cite{dwork2015generalizationarXiv})  for adaptive queries $Q_1,Q_2,\ldots$, each with global sensitivity upper bounded by $\Delta_{Q}$.}
\label{algadaptivedataanalysis}
\begin{algorithmic}[1]
\REQUIRE Holdout dataset $X$, training dataset $D$, noise rate $\sigma$, budget $B$, threshold $T$, queries $Q_1,Q_2,\ldots$
\ENSURE A stream of answers
\STATE
\textbf{let} $\sigma_2 \leftarrow 4\sigma$, and  $\sigma_3 \leftarrow 2\sigma$;
\STATE \textbf{let}  $\widehat{T} \leftarrow T + \textrm{Lap}(\sigma)$;~~~~~\COMMENT{Comment: $\textrm{Lap}(\sigma)$ means sampling a fresh Laplace noise with parameter (i.e., scale) $\sigma$.}
\FOR{each query $Q_{i}$}
\IF{$B<1$}
  \STATE \textbf{output} ``$\perp$'';
\ELSE
\STATE \textbf{let} $\gamma_{i} \leftarrow \textrm{Lap}(\sigma_3)$;
\IF{$|Q_{i}(X)-Q_{i}(D)|+\gamma_{i}>\widehat{T}$}
  \STATE \textbf{let}  $B \leftarrow B-1$ and $\widehat{T} \leftarrow T + \textrm{Lap}(\sigma)$;
 \STATE \textbf{output} $Q_{i}(X)+\textrm{Lap}(\sigma_2)$;
\ELSE
\STATE \textbf{output} $Q_{i}(D)$; 
\ENDIF
\ENDIF
\ENDFOR
\end{algorithmic}
\end{algorithm}

\section{The Results} \label{sec:main:res}



 Our algorithm for adaptive statistical learning with correlated samples is presented as Algorithm \ref{algadaptivedataanalysis} (based on \cite{dwork2015generalizationarXiv}), which tackles adaptive queries $Q_1,Q_2,\ldots$, each with global sensitivity upper bounded by $\Delta_{Q}$. The queries $Q_1,Q_2,\ldots$   are chosen adaptively based on the results of prior estimates.
We use Algorithm \ref{algadaptivedataanalysis} to enable validation of an analyst's queries in the adaptive setting. As will become clear, Algorithm \ref{algadaptivedataanalysis} is $\epsilon$-differential private for \mbox{$\epsilon=B \Delta_{Q} \times \big(\frac{1}{\sigma} \hspace{-1pt} +\hspace{-1pt}\frac{1}{\sigma_2}\hspace{-1pt}+\hspace{-1pt}\frac{2}{\sigma_3}\big)$} $ =\frac{9B \Delta_{Q}}{4\sigma}$   and thus is $\epsilon'$-Bayesian differential private for $\epsilon'$ defined in Lemma \ref{thmDPBDP-main-result-generalized} or Lemma \ref{thmDPBDP-main-result} on Page \pageref{thmDPBDP-main-result-generalized} later with the above $\epsilon$. For an unknown distribution $\mathcal{D}$ over a discrete
universe $\mathcal{X}$ of possible data points, a statistical query
$Q$ asks for the expected value of some function $f: \mathcal{X} \rightarrow [0,1]$ on random draws from $\mathcal{D}$. It follows that $\Delta_Q=1/n$ for a statistical query \cite{dwork2015generalizationarXiv,bassily2016typicality,dwork2015preserving}.

\begin{thm} \label{lem-boundgeneralization-eachi}
Let $\beta,\tau,\sigma,B, T > 0$. Let $X=[X_1,X_2,\ldots,X_n]$ denote the holdout dataset drawn randomly from a distribution $\mathcal{P}$.
Consider an analyst that is given access to the training dataset $D$ and selects statistical queries $Q_1,Q_2,\ldots$ adaptively
while interacting with Algorithm \ref{algadaptivedataanalysis} which is given holdout dataset $X$, training dataset $D$, noise rate $\sigma$, budget $B$, threshold $T$.
 If
 \begin{align}
n  & \geq n_{*}(B,\sigma,\tau,\beta) \nonumber \\ &  \de
\max\bigg\{ \frac{9\ln(4/\beta)}{\tau^2}, \max\limits_{i\in \{1,\iffalse 2,\fi \ldots,n\}}  ~\frac{9B }{4\sigma \cdot (\tau/3-4 a_i)}\bigg\},\label{nstar}
\end{align}
or
\begin{align}
n  & \geq n_{\#}(B,\sigma,\tau,\beta) \nonumber \\ &  \de
\max\bigg\{ \frac{9\ln(4/\beta)}{\tau^2}, \max\limits_{i\in \{1,\iffalse 2,\fi \ldots,n\}} ~\min\limits_{\text{\rm Markov quilt }X_{Q_i}\text{ of } X_i}~\frac{9B \cdot(|N_i|+1) }{4\sigma \cdot (\tau/3-4 b_i)}\bigg\},\label{npound}
\end{align}
then for arbitrary correlations between data samples, we have
\begin{align}
\text{$\bp{|Q_i(X)-Q_i(\mathcal{P})|\geq \tau}\leq \beta$ for every $i=1,2,\ldots,m$,}
\end{align}
where $a_{i} \de \mathcal{I}(X_{M_i}\leftsquigarrow X_i) = \ln \max\limits_{x_i,x_i',x_{M_i}}\frac{\bp{x_{M_i }\con x_i}}{\bp{x_{M_i }\con x_i'}}$ with $X_{\iffalse\mathcal\fi{M}_i}$ being the Markov blanket of $X_i$, and $b_{i} \de \max\limits_{L\subseteq N_i}\mathcal{I}(X_{Q_i}\leftsquigarrow X_i\con X_L) $ for $\mathcal{I}(X_{Q_i}\leftsquigarrow X_i\con X_L)\de \max\limits_{L\subseteq N_i}\ln \max\limits_{x_{Q_i},x_i,x_i',x_L} \frac{\bp{x_{Q_i}\con x_i,x_L}}{\bp{x_{Q_i}\con x_i',x_L}}  $. In the expression of $a_i$, the records $x_i$ and $x_i'$ iterate through the domain of tuple $X_i$ (i.e., $x_i \in \textrm{domain}(X_i)$, $x_i' \in \textrm{domain}(X_i)$, and $x_i \neq x_i' $), and   $x_{M_i }$ iterates through the domain of tuple(s) $X_{M_i }$ with $X_{\iffalse\mathcal\fi{M}_i}$ being the Markov blanket of $X_i$. In $n_{\#}(\cdot)$ of (\ref{npound}), note that we can let $X_{Q_i}$ iterate through just an arbitrary set containing some Markov quilts of $X_i$, rather than iterating through the set of all Markov quilts of $X_i$, since  letting   $X_{Q_i}$ iterate a smaller set can only induce a larger (or the same) bound in (\ref{npound}). In addition, as explained on Page \pageref{defquilt}, when we define a Markov quilt $X_{Q_i}$ for node $X_i$, the \textit{nearby set} $X_{N_i}$ is also determined; i.e.,   $N_i$ is also determined. Given $N_i$, the set $L$ in the definition of $b_i$ iterates all subsets of  $N_i$. In the expression of $\mathcal{I}(X_{Q_i}\leftsquigarrow X_i\con X_L)$, the records $x_i$ and $x_i'$ iterate through the domain of tuple $X_i$ (i.e., $x_i \in \textrm{domain}(X_i)$, $x_i' \in \textrm{domain}(X_i)$, and $x_i \neq x_i' $),     $x_{Q_i }$ iterates through the domain of tuple(s) $X_{Q_i }$, and   $x_{L}$ iterates through the domain of tuple(s) $X_{L}$.


\end{thm}

Theorem \ref{lem-boundgeneralization-eachi} will be proved in Section \ref{secaplem-boundgeneralization-eachi}. We now apply our results above to analyze the entire execution of Algorithm \ref{algadaptivedataanalysis}.

\begin{thm} \label{thm-entire-execution}

Let $\beta,\tau> 0$ and $m \geq B > 0$. Set $\sigma=\frac{(1-c)\tau}{12\ln (4m/\beta)}$ and $T=\frac{(1+c)\tau}{2}$ for an arbitrary constant $c\in(0,1)$. For every $i=1,2,\ldots,m$, let $a_i$ be the answer of
Algorithm \ref{algadaptivedataanalysis} on statistical query $Q_i$, and define the counter of overfitting \mbox{$Z_i \de \big|\{j\leq i:|Q_j(D)-Q_j(\mathcal{P})|\geq c \tau\}\big|$.}
 Then for arbitrary correlations between data samples, we have
 \begin{align}
\bp{\exists i\in \{1,2,\ldots,m\}:\begin{array}{l}Z_i<B\textrm{ and }\\|a_i-Q_i(\mathcal{P})|\geq \tau\end{array}} \leq \beta,\nonumber
\end{align}
  for
\begin{align}
\textstyle{n} & \textstyle{ \geq n_{\#}\big(B,\frac{(1-c)\tau}{12\ln (4m/\beta)},\frac{(1-c)\tau}{4},\frac{\beta}{2m}\big)} \nonumber \\ & \textstyle{=O\Bigg(\begin{array}{l} B \ln \frac{m}{\beta}\cdot \max\limits_{i\in \{1,\iffalse 2,\fi \ldots,n\}} ~\min\limits_{\text{\rm Markov quilt }X_{Q_i}\text{ of } X_i}~\frac{|N_i|+1}{\tau\cdot (\tau/3-4 b_i)}\end{array} \Bigg),} \nonumber
\end{align} where $n_{\#}(\cdot)$ is defined in (\ref{npound}) of Theorem~\ref{lem-boundgeneralization-eachi}. When the correlations between data samples are represented by a time-homogeneous Markov chain that is also aperiodic, irreducible and reversible, the above condition on $n$ becomes $n \geq O\big(\frac{B \ln (1/\tau)}{\tau^2 }\cdot \ln \frac{m}{\beta}\big)$.

\end{thm}
Theorem \ref{thm-entire-execution} will be proved in Section \ref{secapthm-entire-execution}.
Theorems~\ref{lem-boundgeneralization-eachi} and~\ref{thm-entire-execution} above generalize recent results of Dwork \emph{et al}. \cite{dwork2015generalizationarXiv} to the case of correlated data.  Theorem 25 of \cite{dwork2015generalizationarXiv} presents the
sample complexity for i.i.d. data samples as $O\big(\frac{B}{\tau^2 }\times \ln \frac{m}{\beta}\big)$, while  our Theorem \ref{thm-entire-execution} gives the
sample complexity as $O\big(\frac{B \ln (1/\tau)}{\tau^2 }\times \ln \frac{m}{\beta}\big)$ when the correlations between data samples are modeled by a Markov chain, by introducing just a small additional expense (i.e., the factor $ \ln (1/\tau)$).

We present in Theorem \ref{thm-BDP-max-information} below that $\epsilon$-Bayesian differential privacy implies a bound on approximate max-information.

\begin{thm} \label{thm-BDP-max-information}
Let $X=[X_1,X_2, \ldots, X_n]$ be the statistical database under consideration, and $Y$ be an $\epsilon$-Bayesian differential private algorithm. Then for any $\beta>0$, it holds that $I_{\infty}^{\beta}(X;Y(X)) \leq \big(2\epsilon^2n +\epsilon\sqrt{2n \ln(2/\beta) }\,\big)\log e$, where $\log $ means the binary logarithm.
\end{thm}

Theorem \ref{thm-BDP-max-information} will be proved in Appendix \ref{secprfthm-BDP-max-information} on Page \pageref{secprfthm-BDP-max-information}.
Theorem~19 of Dwork \emph{et al}. \cite{dwork2015generalizationarXiv} give a simple bound $\epsilon n  \log e$ that is weaker than that of our Theorem \ref{thm-BDP-max-information}. In Appendix \ref{secapcor-BDP-max-information} on Page \pageref{secapcor-BDP-max-information}, we will use
Theorem \ref{thm-BDP-max-information} above to obtain the following Lemma \ref{cor-BDP-max-information} on generalization bounds.

\begin{lem}\label{cor-BDP-max-information}
Let $X$ be a random
database chosen according to distribution $\mathcal{P}$, and $Y$ be an $\epsilon$-Bayesian differential private algorithm for query $Q$ with global sensitivity $\Delta_Q$. Let $Y(\mathcal{P})$ be the expectation of $Y(X)$. If $\epsilon\leq {\tau}/({3n\Delta_Q})$, then $\bp{|Y(X)-Y(\mathcal{P})|\geq \tau} \leq 4e^{-{\tau^2}/({9n\cdot{\Delta_Q}^2})}$.
\end{lem}

From Lemma \ref{cor-BDP-max-information}, for a statistical query \cite{dwork2015generalizationarXiv,bassily2016typicality,dwork2015preserving} with $\Delta_Q=1/n$, if $\epsilon\leq {\tau}/{3}$, then $\bp{|Y(X)-Y(\mathcal{P})|\geq \tau}\leq 4e^{-{\tau^2n}/{9}}$.

We use Lemma \ref{cor-BDP-max-information} in the proofs of Theorems \ref{lem-boundgeneralization-eachi} and \ref{thm-entire-execution} for quantifying the number of samples needed to bound the generalization error.

\section{Experiments of Adaptive Statistical Learning} \label{sec:Experiments}

\begin{figure}[!t]
\vspace{-10pt}\centerline{\includegraphics[width=0.43\textwidth]{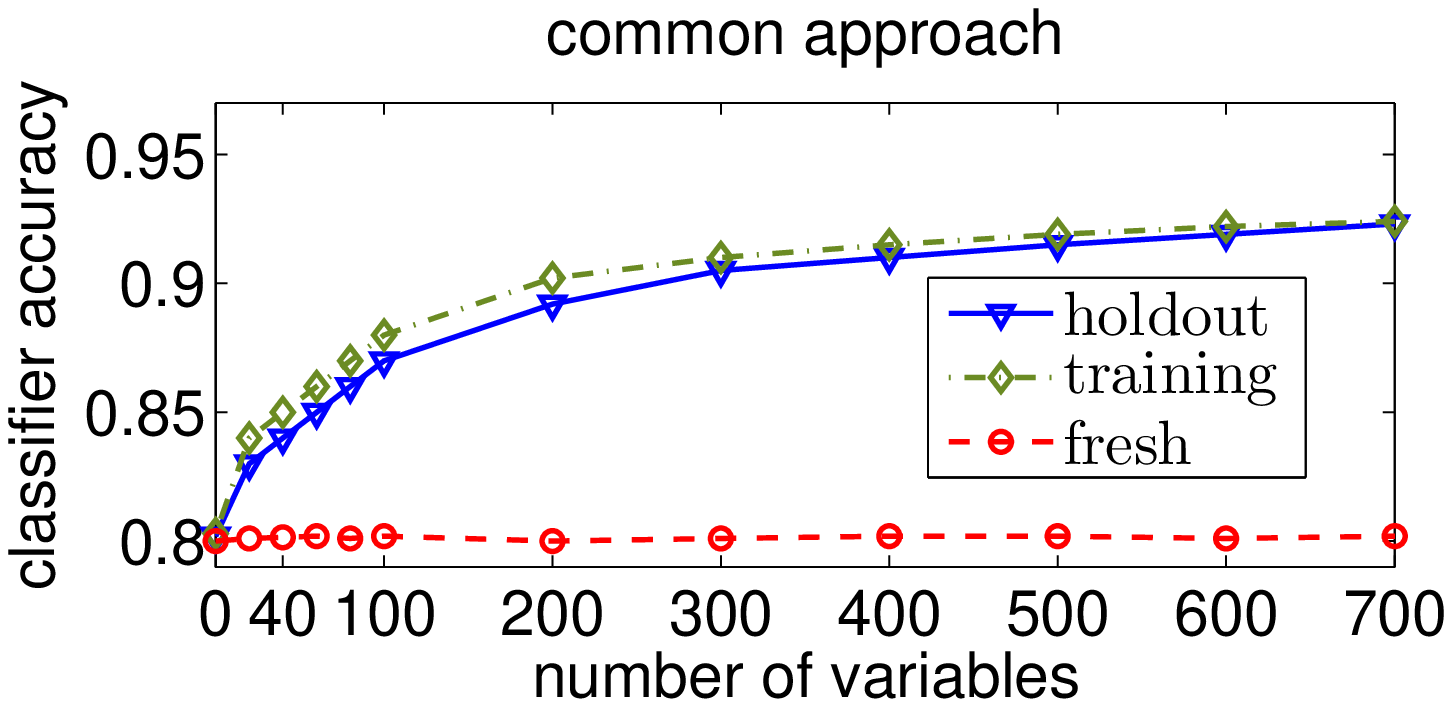}}
\vspace{-10pt}
 \caption{Reusing a holdout
dataset in the common way can lead to overfitting.\vspace{15pt}}
\label{ccs-ada0} \centerline{\includegraphics[width=0.43\textwidth]{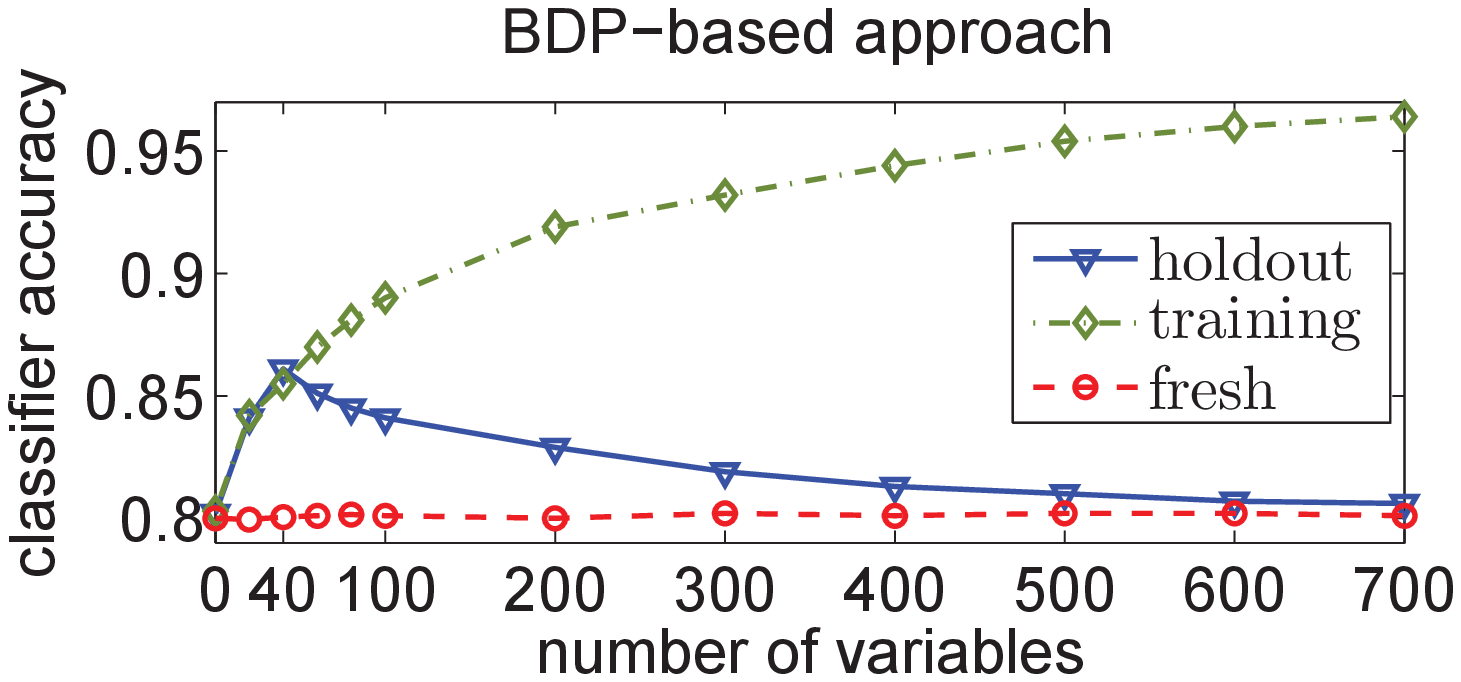}}
\vspace{-10pt}
 \caption{Overfitting of reusing a holdout dataset is prevented in our Bayesian differential privacy-based approach.\vspace{-5pt}}
\label{ccs-ada}
\end{figure}

We provide experiments on synthetic data to support our result of adaptive statistical learning that reuses a holdout
dataset with tuple correlations. The goal of the analyst is
build a linear threshold classifier, which is the same as \cite{dwork2015generalizationarXiv}. We do not repeat the details of the classifier of \cite{dwork2015generalizationarXiv} here. The label is randomly selected from $\{-1,1\}$ and
 the attributes are generated such that they are correlated with the label and also correlated among themselves. Figures \ref{ccs-ada0} and \ref{ccs-ada}
illustrate that reusing a holdout
dataset in the common way can lead to overfitting, and that overfitting is prevented by our approach based on Bayesian differential privacy (BDP).

\section{Proofs}\label{sec:Proofs}

\subsection{Useful Lemmas}\label{sec:Useful}

Below we state several   lemmas that will be used later to prove the theorems. Lemmas \ref{thmDPBDP-main-result-generalized} and \ref{thmDPBDP-main-result} below
present the relationship between Bayesian differential privacy (BDP) and differential privacy (DP) when the tuple correlations can be arbitrary. Lemma \ref{lemexactformoffunctionhdpddp}
provides the corresponding result when   the tuple correlations are modeled by a time-homogeneous Markov chain.

\begin{lem}[\cite{BDPIT}]  \label{thmDPBDP-main-result-generalized} For a database with arbitrary tuple correlations, it holds that
\begin{align}
& \textrm{$\epsilon$-Bayesian differential privacy} \Longleftarrow \textrm{$\epsilon'$-differential privacy} \nonumber \\ & \textrm{for $\epsilon' = \min_{i\in\{1,2,\ldots,n\}} \hspace{6pt}\max_{\begin{subarray}~X_{Q_i}:~\text{\rm a \textit{Markov quilt} of }X_i\\ X_{N_i}:~\text{\rm the \textit{nearby set} associated with }X_{Q_i}\end{subarray}}\hspace{3pt}\frac{\epsilon-4b_{i}}{|N_i|+1}$}, \label{BDP-DP-IMiKL1-generalized}
\end{align}
where  $b_{i} \de \max\limits_{L\subseteq N_i}\mathcal{I}(X_{Q_i}\leftsquigarrow X_i\con X_L)$, the meanings of a Markov quilt and its associated nearby set have been elaborated on Page \pageref{def-Markov-quilt-formal}, and $X_{Q_i}$ in (\ref{BDP-DP-IMiKL1-generalized}) can iterate through the set of all Markov quilts of $X_i$, or just an arbitrary set containing some Markov quilts of $X_i$. Note that we can let $X_{Q_i}$ iterate through just an arbitrary set containing some Markov quilts of $X_i$, rather than iterating through the set of all Markov quilts of $X_i$, since (i) based on (\ref{BDP-DP-IMiKL1-generalized}), letting   $X_{Q_i}$ iterate a smaller set cannot make $\epsilon'$ larger (i.e., it either induces a smaller $\epsilon'$ or does not change $\epsilon'$), and (ii) $\epsilon_*$-differential privacy with a smaller $\epsilon_*$ implies $\epsilon_{\#}$-differential privacy with a larger~$\epsilon_{\#}$. In addition, as explained on Page \pageref{defquilt}, when we define a Markov quilt $X_{Q_i}$ for node $X_i$, the \textit{nearby set} $X_{N_i}$ is also determined; i.e.,   $N_i$ is also determined. Given $N_i$, the set $L$ in the definition of $b_i$ iterates all subsets of  $N_i$.
 \end{lem}

\begin{lem}[\cite{BDPIT}]  \label{thmDPBDP-main-result}
Letting $X_{Q_i}$ in (\ref{BDP-DP-IMiKL1-generalized}) take just the   Markov blanket of $X_i$ since the   Markov blanket is a special case of the   Markov quilt as discussed on Page \pageref{defquilt}, we can replace (\ref{BDP-DP-IMiKL1-generalized}) by \begin{align}
&   \textrm{$\epsilon$-Bayesian differential privacy} \Longleftarrow \textrm{$\epsilon'$-differential privacy} \nonumber \\ & \textrm{for $\epsilon' = \min_{i\in\{1,2,\ldots,n\}} \hspace{6pt} (\epsilon-4a_{i})$}, \nonumber 
\end{align}
where $a_{i} \de \mathcal{I}(X_{M_i}\leftsquigarrow X_i) = \ln \max\limits_{x_i,x_i',x_{M_i}}\frac{\bp{x_{M_i }\con x_i}}{\bp{x_{M_i }\con x_i'}}$.
 \end{lem}


\begin{lem}[\cite{BDPIT}] \label{lemexactformoffunctionhdpddp}
Consider a database with tuples modeled by a time-homogeneous Markov chain $X_1 \rightarrow X_2 \rightarrow \ldots \rightarrow X_n$ that is also aperiodic, irreducible and reversible. For this Markov chain, let $g$ be the spectral gap of the transition matrix; i.e., $g$ equals $1-\max\{|\lambda_2|,|\lambda_3|,\ldots,|\lambda_n|\}$ with the eigenvalues of the transition matrix in non-increasing order being $\lambda_1\geq \lambda_2\geq \ldots\geq \lambda_n$, where $\lambda_1=1$. Let  $\rho$ be the probability of the least probable state in the stationary
distribution of the Markov chain; i.e., $\rho$ equals $\min_{j\in S}d_j$ with vector $[d_j:j\in S]$ denoting the stationary distribution and $S$ denoting the state space. Let $c$ be an arbitrary constant satisfying $0<c<1/6$. With $ d \de \Big\lceil\frac{1}{g}\ln \frac{e^{c\epsilon}+1}{\rho(e^{c\epsilon}-1)}\Big\rceil$, $t \de \big\lfloor\frac{1}{g}\ln \frac{1}{\rho}\big\rfloor$ and $\xi_{t+1} \de \frac{e^{-g(t+1)}}{\rho}$, then for $n \geq 2d$, $\epsilon$-Bayesian differential privacy is implied by $h(\epsilon,g,\rho)$-differential privacy, where
\begin{align}
h(\epsilon,g,\rho)  & \de  \min\Bigg\{\frac{(1-6c)\epsilon}{2d-1},~\frac{(\frac{1}{3}-2c)\epsilon}{d+s} \Bigg\}.\label{hepsgrhoimproved2}
\end{align}
\end{lem}

\def\x{\ref{lem-boundgeneralization-eachi}}
\def\y{\pageref{lem-boundgeneralization-eachi}}

\subsection{Proof of Theorem~\x~on Page \y} \label{secaplem-boundgeneralization-eachi}

For an unknown distribution $\mathcal{D}$ over a discrete
universe $\mathcal{X}$ of possible data points, a statistical query
$Q$ asks for the expected value of some function $f: \mathcal{X} \rightarrow [0,1]$ on random draws from $\mathcal{D}$. On a database of $n$ records, the global sensitivity of a statistical query
$Q$ is $\Delta_Q=1/n$. Substituting $\Delta_Q=1/n$ into Lemma~\ref{cor-BDP-max-information} on Page~\pageref{cor-BDP-max-information}, we obtain that under $\epsilon$-Bayesian differential privacy, if $\epsilon\leq {\tau}/3$, then \mbox{$\bp{|Y(X)-Y(\mathcal{P})|\geq \tau} \leq 4e^{-{\tau^2}n/9}$.}  Hence, to induce the desired $\bp{|Q_i(X)-Q_i(\mathcal{P})|\geq \tau}\leq \beta$ in Theorem~\ref{lem-boundgeneralization-eachi}, we ensure $\epsilon\leq {\tau}/3$ and $4e^{-{\tau^2}n/9}\leq \beta$. From \cite{dwork2015generalizationarXiv,dwork2014algorithmic}, Algorithm \ref{algadaptivedataanalysis} on Page~\pageref{algadaptivedataanalysis} is $\epsilon_*$-differential private for $\epsilon_*=B \Delta_{Q} \cdot \big(\frac{1}{\sigma}+\frac{1}{\sigma_2}+\frac{2}{\sigma_3}\big)=\frac{9B \Delta_{Q}}{4\sigma}$. The above result and Lemma \ref{thmDPBDP-main-result-generalized} together imply that Algorithm \ref{algadaptivedataanalysis} is $\epsilon$-Bayesian differential private for $\epsilon\hspace{-1pt} =\hspace{-1pt} \min\limits_{i\in \{1,\iffalse 2,\fi \ldots,n\}}\hspace{-1pt} ~\max\limits_{\text{\rm Markov quilt }X_{Q_i}\text{ of } X_i}\hspace{-1pt} \frac{\frac{9B \Delta_{Q}}{4\sigma}-4b_i}{|N_i|+1}$. Then $\Delta_Q=1/n$ and the condition $\epsilon\leq {\tau}/3$ imply
\begin{align}
\textstyle{n \geq  \max\limits_{i\in \{1,\iffalse 2,\fi \ldots,n\}} ~\min\limits_{\text{\rm Markov quilt }X_{Q_i}\text{ of } X_i}\frac{9B \cdot(|N_i|+1) }{4\sigma \cdot (\tau/3-4 b_i)} }. \label{nboundlem1}
\end{align}
 In addition, the condition $4e^{-{\tau^2}n/9}\leq \beta$ implies
\begin{align}
\textstyle{n \geq \frac{9\ln(4/\beta)}{\tau^2}}. \label{nboundlem2}
\end{align}
Then with $n_{\#}(B,\sigma,\tau,\beta)$ defined as the maximum of the right hand sides of (\ref{nboundlem1}) and (\ref{nboundlem2}), we obtain that $n \geq n_{\#}(B,\sigma,\tau,\beta) $ suffices.
 We can also apply Lemma \ref{thmDPBDP-main-result} instead of Lemma \ref{thmDPBDP-main-result-generalized}. In this case, (\ref{nboundlem1}) is replaced by
\begin{align}
\textstyle{n \geq  \max\limits_{i\in \{1,\iffalse 2,\fi \ldots,n\}} \frac{9B   }{4\sigma \cdot (\tau/3-4 a_i)} }, \nonumber
\end{align}
  which together with (\ref{nboundlem2}) implies that $n \geq n_{*}(B,\sigma,\tau,\beta) $ suffices. \pfe

\def\x{\ref{thm-entire-execution}}
\def\y{\pageref{thm-entire-execution}}
\subsection{Proof of Theorem~\x~on Page \y} \label{secapthm-entire-execution}

We prove Theorem \ref{thm-entire-execution} using the technique similar to that of \cite[Theorem 25]{dwork2015generalizationarXiv}.
For notation convenience in the analysis, we write ``$\widehat{T} \leftarrow T + \textrm{Lap}(\sigma)$'' in Lines 2 and 9 of Algorithm \ref{algadaptivedataanalysis} as ``$\widehat{T} \leftarrow T + \delta_i$'', where $ \delta_i$ denotes $\textrm{Lap}(\sigma)$; i.e., $ \delta_i$ is a fresh Laplace noise with parameter (i.e., scale) $\sigma$. Also, we write $Q_i(X)+\textrm{Lap}(\sigma_2)$ in Line 10 of Algorithm \ref{algadaptivedataanalysis} as $Q_i(X)+ \xi_i$, where $ \xi_i$ denotes $\textrm{Lap}(\sigma_2)$; i.e., $ \xi_i$ is a fresh Laplace noise with parameter $\sigma_2$. With the above changes, we restate Algorithm~\ref{algadaptivedataanalysis}.
\setcounter{algorithm}{0}
\begin{algorithm}
\caption{\textbf{(Restated).} A generalization-achieving algorithm (based on \cite{dwork2015generalizationarXiv}) for adaptive queries $Q_1,Q_2,\ldots$, each with global sensitivity upper bounded by $\Delta_{Q}$.}
\label{algadaptivedataanalysisnew}
\begin{algorithmic}[1]
\REQUIRE Holdout set $X$, training set $D$, noise rate $\sigma$, budget $B$, threshold $T$, queries $Q_1,Q_2,\ldots$
\ENSURE A stream of answers
\STATE
\textbf{let} $\sigma_2 \leftarrow 4\sigma$, and  $\sigma_3 \leftarrow 2\sigma$;
\STATE \textbf{let}  $\widehat{T} \leftarrow T + \delta_i$, where $ \delta_i \leftarrow \textrm{Lap}(\sigma)$;~~~~~\COMMENT{Comment: $\textrm{Lap}(\sigma)$ means sampling a fresh Laplace noise with parameter (i.e., scale) $\sigma$.}
\FOR{each query $Q_{i}$}
\IF{$B<1$}
  \STATE \textbf{output} ``$\perp$'';
\ELSE
\STATE \textbf{let} $ \gamma_{i} \leftarrow \textrm{Lap}(\sigma_3)$;
\IF{$|Q_{i}(X)-Q_{i}(D)|+ \gamma_{i}>\widehat{T}$}
  \STATE \textbf{let}  $B \leftarrow B-1$ and $\widehat{T} \leftarrow T + \delta_i$, where $ \delta_i \leftarrow \textrm{Lap}(\sigma)$;
 \STATE \textbf{output} $Q_{i}(X)+ \xi_i$, where $ \xi_i\leftarrow  \textrm{Lap}(\sigma_2)$;
\ELSE
\STATE \textbf{output} $Q_{i}(D)$; 
\ENDIF
\ENDIF
\ENDFOR
\end{algorithmic}
\end{algorithm}

Recall that $a_i$ denotes the answer of
Algorithm \ref{algadaptivedataanalysis} on statistical query $Q_i$. In the result that we desire to prove, we bound the error between $a_i$ and $Q_i(\mathcal{P})$. This error can be decomposed as the difference between $a_i$ and $Q_i(X)$, and the difference between $Q_i(X)$ and $Q_i(\mathcal{P})$. More specifically, \h
\begin{align}
& \bp{a_i \neq \perp \& |a_i-Q_i(\mathcal{P})|\geq \tau} \nonumber  \\ &\hspace{-2pt} = \hspace{-2pt}\bp{{a_i \hspace{-2pt}\neq \hspace{-1pt}\perp \&|a_i\hspace{-2pt}-\hspace{-2pt}Q_i(X)|\hspace{-2pt}\geq\hspace{-2pt} \frac{3+c}{4}\tau}}\hspace{-2pt}+\hspace{-2pt}\bp{{|Q_i(\mathcal{P})\hspace{-2pt}-\hspace{-2pt}Q_i(X)|\hspace{-2pt}\geq \hspace{-2pt}\frac{1-c}{4}\tau}} . \label{decomposeerror}
\end{align}
A more formal reasoning that uses the union bound to establish (\ref{decomposeerror}) is presented below, where $\overline{A}$ means the complement of event~$A$. We have
{
\begin{align}
& \bp{a_i \neq \perp \& |a_i-Q_i(\mathcal{P})|\geq \tau} \nonumber \\ &\hspace{-2pt}=\hspace{-2pt} \bp{\overline{a_i =\perp ~\text{or}~ |a_i-Q_i(\mathcal{P})|< \tau}} \nonumber \\ & \hspace{-2pt}\leq\hspace{-2pt} \bp{\overline{\bigg(a_i \hspace{-2pt}= \hspace{-1pt}\perp\text{or}\hspace{2pt}|a_i\hspace{-2pt}-\hspace{-2pt}Q_i(X)|\hspace{-2pt}<\hspace{-2pt} \frac{3+c}{4} \tau\bigg) \& \bigg(|Q_i(\mathcal{P})\hspace{-2pt}-\hspace{-2pt}Q_i(X)|\hspace{-2pt}<\hspace{-2pt} \frac{1-c}{4} \tau\bigg)}} \nonumber \\ & \hspace{-2pt}= \hspace{-2pt}\bp{\overline{a_i \hspace{-2pt}= \hspace{-1pt}\perp \text{or}\hspace{2pt}|a_i\hspace{-2pt}-\hspace{-2pt}Q_i(X)|\hspace{-2pt}<\hspace{-2pt} \frac{3+c}{4}\tau}~\text{or}~\overline{|Q_i(\mathcal{P})\hspace{-2pt}-\hspace{-2pt}Q_i(X)|\hspace{-2pt}< \hspace{-2pt}\frac{1-c}{4}\tau}}  \nonumber \\ &\hspace{-2pt} \leq\hspace{-2pt} \bp{\overline{a_i \hspace{-2pt}= \hspace{-1pt}\perp \text{or}\hspace{2pt}|a_i\hspace{-2pt}-\hspace{-2pt}Q_i(X)|\hspace{-2pt}<\hspace{-2pt} \frac{3+c}{4}\tau}}\hspace{-2pt}+\hspace{-2pt}\bp{\overline{|Q_i(\mathcal{P})\hspace{-2pt}-\hspace{-2pt}Q_i(X)|\hspace{-2pt}< \hspace{-2pt}\frac{1-c}{4}\tau}}   \nonumber  \\ &\hspace{-2pt} = \hspace{-2pt}\bp{{a_i \hspace{-2pt}\neq \hspace{-1pt}\perp \&|a_i\hspace{-2pt}-\hspace{-2pt}Q_i(X)|\hspace{-2pt}\geq\hspace{-2pt} \frac{3+c}{4}\tau}}\hspace{-2pt}+\hspace{-2pt}\bp{{|Q_i(\mathcal{P})\hspace{-2pt}-\hspace{-2pt}Q_i(X)|\hspace{-2pt}\geq \hspace{-2pt}\frac{1-c}{4}\tau}} .  \label{decomposeerror2}
\end{align}}
To bound the second term in (\ref{decomposeerror2}), we use Theorem~\ref{lem-boundgeneralization-eachi} to obtain
\begin{align}
 & \text{for every }i:\bp{{|Q_i(\mathcal{P})-Q_i(X)|\geq \frac{1-c}{4}\tau}}  \leq \frac{\beta}{2m} \label{foreveryiQiPX}\\ & \text{ if }n  \geq n_{\#}\bigg(B,\sigma,\frac{(1-c)\tau}{4},\frac{\beta}{2m}\bigg). \nonumber
\end{align}
where ``for every $i$'' means ``$i \in \{1,\ldots,m\}$'' (note that $m$ is the number of queries answered). Then (\ref{foreveryiQiPX}) and the union bound together imply
\begin{align}
 &  \bp{\exists i:{|Q_i(\mathcal{P})-Q_i(X)|\geq \frac{1-c}{4}\tau}}  \leq \frac{\beta}{2} \label{foreveryiQiPXabc}\\ & \text{ if }n  \geq n_{\#}\bigg(B,\sigma,\frac{(1-c)\tau}{4},\frac{\beta}{2m}\bigg), \nonumber
\end{align}
where ``$\exists i$'' means ``$\exists i \in \{1,\ldots,m\}$''.

Below we bound the first term in (\ref{decomposeerror2}) by analyzing Algorithm \ref{algadaptivedataanalysis}.
For the answer $a_i$ that is different from $\perp$, we bound the first term in (\ref{decomposeerror2}) by considering two cases of Algorithm \ref{algadaptivedataanalysis}. First, if Line 10 of Algorithm \ref{algadaptivedataanalysis} is executed, then $a_i=Q_{i}(X)+ \xi_i$ and thus $|a_i-Q_{i}(X)|=| \xi_i|$. Second, if Line 12 of Algorithm \ref{algadaptivedataanalysis} is executed, then $|Q_{i}(X)-Q_{i}(D)|+ \gamma_{i}\leq\widehat{T}$ and $a_i=Q_{i}(D)$, yielding $|a_i-Q_{i}(X)|=|Q_{i}(D)-Q_{i}(X)|\leq \widehat{T}- \gamma_{i}=T+ \delta_i -  \gamma_{i}$ and furthermore $|a_i-Q_{i}(X)| \leq  T+| \delta_i|+| \gamma_{i}|$. Summarizing the two cases above, we obtain
{
\begin{align}
 & \bp{\exists i:{a_i \neq \perp ~\&~|a_i-Q_i(X)|\geq \frac{3+c}{4}\cdot\tau}}\nonumber \\ & \leq \max\bigg\{\bp{\exists i:| \xi_i|\geq \frac{3+c}{4} \tau},\bp{\exists i:| \delta_i|+| \gamma_{i}|\geq \frac{3+c}{4} \tau-T}\hspace{-2pt}\bigg\}\nonumber \\ & \leq \max\bigg\{\bp{\exists i:| \xi_i|\geq \frac{3+c}{4} \tau},\bp{\exists i:| \delta_i|+| \gamma_{i}|\geq \frac{1-c}{4} \tau}\hspace{-2pt}\bigg\}, \label{probexitaiqix}
\end{align}}
where the last step uses $T=\frac{(1+c)\tau}{2}$.

Since $ \xi_{i}$ obeys a Laplace distribution with parameter $4\sigma$, where $\sigma \de \frac{(1-c)\tau}{12\ln (4m/\beta)}$ from the condition, \h
\begin{align}
\text{for every }i:\bp{| \xi_i|\geq \frac{3+c}{4} \tau} = \exp\bigg\{-\frac{\frac{3+c}{4} \cdot\tau}{4\sigma}\bigg\} \leq \frac{\beta}{4m}, \label{boundxii}
\end{align}
where the last step uses $\frac{3+c}{16} \geq \frac{1-c}{12}$. Then (\ref{boundxii}) and the union bound    yield
\begin{align}
 \bp{\exists i:| \xi_i|\geq \frac{3+c}{4} \tau}   \leq \frac{\beta}{4}. \label{boundxii2}
\end{align}

To bound $\bp{\exists i:| \delta_i|+| \gamma_{i}|\geq \frac{1-c}{4} \tau}$, we use the union bound to derive
\begin{align}
 & \bp{\exists i:| \delta_i|+| \gamma_{i}|\geq \frac{1-c}{4} \tau}\nonumber \\ & =  \bp{\overline{\forall i:| \delta_i|+| \gamma_{i}|< \frac{1-c}{4} \tau}}\nonumber \\ & \leq \bp{\overline{\bigg(\forall i:| \delta_i|< \frac{1-c}{12} \tau\bigg)~\&~\bigg(\forall i:| \gamma_i|< \frac{1-c}{6} \tau\bigg)}} \nonumber \\ & = \bp{\overline{\bigg(\forall i:| \delta_i|< \frac{1-c}{12} \tau\bigg)} \text{ or }\overline{\bigg(\forall i:| \gamma_i|< \frac{1-c}{6} \tau\bigg)}}\nonumber \\ & \leq \bp{\overline{\bigg(\forall i:| \delta_i|< \frac{1-c}{12} \tau\bigg)}}+ \bp{\overline{\bigg(\forall i:| \gamma_i|< \frac{1-c}{6} \tau\bigg)}}\nonumber \\ & = \bp{\exists i:| \delta_i| \geq \frac{1-c}{12} \tau} + \bp{\exists i: | \gamma_{i}|\geq \frac{1-c}{6} \tau}. \label{boundexistgammaeta}
\end{align}

Since $ \delta_i$ obeys a Laplace distribution with parameter $\sigma$, and $ \gamma_i$ obeys a Laplace distribution with parameter $2\sigma$, then given $\sigma=\frac{(1-c)\tau}{12\ln (4m/\beta)}$, we obtain
\begin{align}
\text{for every }i:\bp{| \delta_i|\geq \frac{1-c}{12}\cdot\tau} =\exp\bigg\{-\frac{\frac{1-c}{12}\cdot\tau}{\sigma}\bigg\} = \frac{\beta}{4m} \label{boundgammai}
\end{align}
and
\begin{align}
\text{for every }i:\bp{| \gamma_i|\geq \frac{1-c}{6}\cdot\tau} = \exp\bigg\{-\frac{\frac{1-c}{6}\cdot\tau}{2\sigma}\bigg\} = \frac{\beta}{4m}. \label{boundetai}
\end{align}
Then using the union bound together with
(\ref{boundgammai}) and (\ref{boundetai}), we have
\begin{align}
\bp{\exists i:| \delta_i|\geq \frac{1-c}{12}\cdot\tau} =\exp\bigg\{-\frac{\frac{1-c}{12}\cdot\tau}{\sigma}\bigg\} = \frac{\beta}{4} \label{boundgammai2}
\end{align}
and
\begin{align}
\bp{\exists i:| \gamma_i|\geq \frac{1-c}{6}\cdot\tau} = \exp\bigg\{-\frac{\frac{1-c}{6}\cdot\tau}{2\sigma}\bigg\} = \frac{\beta}{4}. \label{boundetai2}
\end{align}
Applying (\ref{boundgammai2}) and (\ref{boundetai2}) to (\ref{boundexistgammaeta}), we derive
\begin{align}
\bp{\exists i:| \delta_i|+| \gamma_{i}|\geq \frac{1-c}{4} \tau}  &  \leq \frac{\beta}{2}.
 \label{boundetai2align}
\end{align}
Substituting (\ref{boundxii2}) and (\ref{boundetai2align}) into (\ref{probexitaiqix}), we obtain
\begin{align}
 \bp{\exists i:{a_i \neq \perp ~\&~|a_i-Q_i(X)|\geq \frac{3+c}{4}\cdot\tau}}  \leq \frac{\beta}{2}.
 \label{boundetai2align2}
\end{align}

Using (\ref{foreveryiQiPXabc}) and (\ref{boundetai2align2}) in (\ref{decomposeerror2}), we have
\begin{align}
&  \bp{\exists i:a_i \neq \perp ~\&~ |a_i-Q_i(\mathcal{P})|\geq \tau}  \leq \beta. \label{decomposeerrorsdz} \\ & \text{ if }n  \geq n_{\#}\bigg(B,\sigma,\frac{(1-c)\tau}{4},\frac{\beta}{2m}\bigg). \nonumber
\end{align}
To complete the proof, we will show that $Z_i<B$ is a subevent of $a_i \neq \perp$ (i.e., if $Z_i<B$ then $a_i \neq \perp$) so that
\begin{align}
 & \bp{\exists i:Z_i<B ~\&~ |a_i-Q_i(\mathcal{P})|\geq \tau} \nonumber \\ & \leq \bp{\exists i:a_i \neq \perp ~\&~ |a_i-Q_i(\mathcal{P})|\geq \tau}, \nonumber
\end{align} where
$Z_i \de \big|\{j\leq i:|Q_j(D)-Q_j(\mathcal{P})|\geq c \tau\}\big|$ denotes the counter of overfitting. For every $j \leq i$ that reduces the budget $B$ in Algorithm \ref{algadaptivedataanalysis} (i.e., Line 9 of  Algorithm \ref{algadaptivedataanalysis} is executed), \f
\begin{align}
  |Q_j(D)-Q_j(\mathcal{P})|&\geq |Q_j(X)-Q_j(D)| - |Q_j(\mathcal{P})-Q_j(X)| \nonumber \\ & \geq |\widehat{T}- \gamma_j|-|Q_j(\mathcal{P})-Q_j(X)| \nonumber \\ & \geq |T+ \delta_j- \gamma_j|-|Q_j(\mathcal{P})-Q_j(X)|  \nonumber \\ & \geq T-| \delta_j|-| \gamma_j|-|Q_j(\mathcal{P})-Q_j(X)| .  \label{Tgammaietaj}
\end{align}
Then applying $T=\frac{1+c}{2} \tau$, $| \delta_j| < \frac{1-c}{12} \tau$, $| \gamma_j|<\frac{1-c}{6} \tau $ and
$|Q_j(\mathcal{P})-Q_j(X)|<\frac{1-c}{4} \tau$ to (\ref{Tgammaietaj}), we find
\begin{align}
  |Q_j(D)-Q_j(\mathcal{P})|&\geq \frac{1+c}{2} \tau- \frac{1-c}{12} \tau- \frac{1-c}{6} \tau  -\frac{1-c}{4} \tau  = c \tau.\nonumber
\end{align}
Hence, if $Z_i<B$, then the budget $B$ in Algorithm \ref{algadaptivedataanalysis} is still at least $1$, and hence $a_i \neq \perp$. This along with (\ref{decomposeerrorsdz}) implies that
\begin{align}
&  \bp{\exists i:a_i \neq \perp ~\&~ |a_i-Q_i(\mathcal{P})|\geq \tau}  \leq \beta. \nonumber \\ & \text{ if }n  \geq n_{\#}\bigg(B,\sigma,\frac{(1-c)\tau}{4},\frac{\beta}{2m}\bigg).  \label{Tgammaietajboundn}
\end{align}
Given the condition $\sigma=\frac{(1-c)\tau}{12\ln (4m/\beta)}$ and $n_{\#}(\cdot)$ defined in Theorem~\ref{lem-boundgeneralization-eachi}, we obtain from (\ref{Tgammaietajboundn}) that
\begin{align}
 n &  \geq n_{\#}\bigg(B,\frac{(1-c)\tau}{12\ln (4m/\beta)},\frac{(1-c)\tau}{4},\frac{\beta}{2m}\bigg)\nonumber \\ & = \max\left\{\begin{array}{l} \frac{9\ln(4/\beta)}{\tau^2}, \\ \max\limits_{i\in \{1,\iffalse 2,\fi \ldots,n\}}  ~\min\limits_{\text{\rm Markov quilt }X_{Q_i}\text{ of } X_i} ~\frac{9B \cdot(|N_i|+1) }{4\cdot \frac{(1-c)\tau}{12\ln (4m/\beta)} \cdot (\tau/3-4 b_i)}\end{array}\right\}\nonumber \\ &= \textstyle{O\Bigg(\begin{array}{l} B \ln \frac{m}{\beta}\cdot \max\limits_{i\in \{1,\iffalse 2,\fi \ldots,n\}} ~\min\limits_{\text{\rm Markov quilt }X_{Q_i}\text{ of } X_i} ~\frac{|N_i|+1}{\tau\cdot (\tau/3-4 b_i)}\end{array} \Bigg)}.  \label{Tgammaietajboundn2}
\end{align}
Hence, we have proved the result for arbitrary   correlations between data samples in Theorem \ref{thm-entire-execution}. Now we establish the result of in Theorem \ref{thm-entire-execution} when the Markov chain represents the  correlations between data samples:
on a time-homogeneous Markov chain that is also aperiodic, irreducible and reversible, the  condition on $n$ in (\ref{Tgammaietajboundn2}) becomes $n = O\big(\frac{B \ln (1/\tau)}{\tau^2 }\cdot \ln \frac{m}{\beta}\big)$. To this end, given (\ref{Tgammaietajboundn2}), then with $\epsilon$ being $\tau/3$, we will evaluate
\begin{align}
\textstyle{\max\limits_{i\in \{1,\iffalse 2,\fi \ldots,n\}} ~\min\limits_{\text{\rm Markov quilt }X_{Q_i}\text{ of } X_i} \frac{|N_i|+1}{  \epsilon-4 b_i}}.\label{epstau3da}
\end{align}
  As shown in Lemma \ref{thmDPBDP-main-result-generalized} on Page \pageref{thmDPBDP-main-result-generalized} above, $\epsilon$-Bayesian differential privacy is implied by $\epsilon'$-differential privacy, for \\$\epsilon'\de \min\limits_{i\in \{1,\iffalse 2,\fi \ldots,n\}}  ~\max\limits_{\text{\rm Markov quilt }X_{Q_i}\text{ of } X_i} \frac{\epsilon-4 b_i}{|N_i|+1}$. Then the term in (\ref{epstau3da}) equals ${1}/{\epsilon'}$. The above result holds because for any sequence $s_i$, we can write $\frac{1}{\max s_i}$ as $\min\frac{1}{ s_i}$, and  write $\frac{1}{\min s_i}$ as $\max\frac{1}{ s_i}$.
   On a time-homogeneous Markov chain $X_1 \rightarrow X_2 \rightarrow \ldots \rightarrow X_n$ that is also aperiodic, irreducible and reversible, the quantity $\rho>0$ is the probability of the least probable state in the stationary
distribution of the Markov chain; i.e., $\rho$ equals $\min_{j\in S}d_j$ with vector $[d_j:j\in S]$ denoting the stationary distribution and~$S$ denoting the state space. The term $g$ is the spectral gap of the transition matrix; i.e., $g$ equals $1-\max\{|\lambda_2|,|\lambda_3|,\ldots,|\lambda_n|\}$ with the eigenvalues of the transition matrix in non-increasing order being $\lambda_1\geq \lambda_2\geq \ldots\geq \lambda_n$, where $\lambda_1=1$. Then Lemma \ref{lemexactformoffunctionhdpddp} on Page \pageref{lemexactformoffunctionhdpddp} above shows that ${\epsilon'}$ above can be replaced by $h(\epsilon,g,\rho)$ as follows:
\begin{align}
h(\epsilon,g,\rho)  & \de  \min\Bigg\{\frac{(1-6c)\epsilon}{2d-1},\frac{(\frac{1}{3}-2c)\epsilon}{d+s} \Bigg\}, \label{hepsgrhoimproved2rec}
\end{align}
where $d \de \Big\lceil\frac{1}{g}\ln \frac{e^{c\epsilon}+1}{\rho(e^{c\epsilon}-1)}\Big\rceil$   and  $s \de \Big\lfloor \frac{1}{g}\ln \frac{e^{\epsilon/6}+1}{\rho(e^{\epsilon/6}-1)}\Big\rfloor$.
 We further  prove that $1/[h(\epsilon,g,\rho)]=O\big(\frac{1}{\tau}\ln \frac{1}{\tau}\big)$ follows. Clearly, we obtain
\begin{align}
d \leq \frac{1}{g}\ln \frac{e^{c\epsilon}+1}{\rho(e^{c\epsilon}-1)} + 1 = \frac{1}{g} \ln\bigg(1+\frac{2}{e^{c\epsilon}-1}\bigg) + \frac{1}{g} \ln \frac{1}{\rho} + 1 \label{ddefecepsas}
\end{align}
and
\begin{align}
s \leq \frac{1}{g}\ln \frac{e^{\epsilon/6}+1}{\rho(e^{\epsilon/6}-1)} = \frac{1}{g}\ln\bigg(1+\frac{2}{e^{\epsilon/6}-1}\bigg) + \frac{1}{g} \ln \frac{1}{\rho}. \label{sdefecepsas}
\end{align}
Given $\epsilon = \tau/3$, $0<\tau \leq 1$, $0<c<1/6$, we have
$c\epsilon < \epsilon/6 \leq \tau/18 \leq 1/18$, implying
$e^{c\epsilon}  < e^{\epsilon/6} \leq e^{1/18}$. Then it holds that
$\frac{2}{e^{c\epsilon}-1} > \frac{2}{e^{\epsilon/6}-1} \geq \frac{2}{e^{1/18}-1} >1  $.
 Since $\ln (1+a) \leq \ln (2a)$ holds for  $a \geq 1$, we further obtain $\ln(1+\frac{2}{e^{c\epsilon}-1}) \leq  \ln \frac{4}{e^{c\epsilon}-1} \leq \ln \frac{4}{c\epsilon} = \ln \frac{12}{c\tau}  $ and $\ln(1+\frac{2}{e^{\epsilon/6}-1}) \leq  \ln \frac{4}{e^{\epsilon/6}-1} \leq \ln \frac{4}{\epsilon/6} = \ln \frac{24}{\epsilon} = \ln \frac{72}{\tau} $. The application of these to (\ref{ddefecepsas}) and (\ref{sdefecepsas}) yields
 \begin{align}
d \leq  \frac{1}{g} \ln \frac{12}{c\tau}  + \frac{1}{g} \ln \frac{1}{\rho}+ 1 \label{ddefecepsas7}
\end{align}
and
\begin{align}
s \leq  \frac{1}{g}\ln \frac{72}{\tau} + \frac{1}{g} \ln \frac{1}{\rho}. \label{sdefecepsas7}
\end{align}
Using (\ref{ddefecepsas7}) and (\ref{sdefecepsas7}) in (\ref{hepsgrhoimproved2rec}), we get
\begin{align}
h(\epsilon,g,\rho)  & \de  \min\Bigg\{\frac{(1-6c)\epsilon}{2d-1},\frac{(\frac{1}{3}-2c)\epsilon}{d+s} \Bigg\} \nonumber \\ & \geq   \min\Bigg\{\frac{(1-6c)\cdot \tau/3}{\frac{2}{g} \ln \frac{12}{c\rho} + \frac{2}{g}\ln \frac{1}{\tau} +1 },\nonumber \\ & \hspace{30pt} \frac{(\frac{1}{3}-2c)\cdot \tau/3}{\frac{1}{g} \ln \frac{12}{c\rho} + \frac{1}{g}\ln \frac{1}{\tau} +1 +\frac{1}{g} \ln \frac{72}{\rho} + \frac{1}{g}\ln \frac{1}{\tau}} \Bigg\},\nonumber
\end{align}
which further implies
\begin{align}
\frac{1}{h(\epsilon,g,\rho) }
 & \leq   \min\Bigg\{
 \frac{\frac{2}{g} \ln \frac{12}{c\rho} + \frac{2}{g}\ln \frac{1}{\tau} +1 }{(1-6c)\cdot \tau/3},\nonumber \\ &\hspace{10pt}
 \frac{\frac{1}{g} \ln \frac{12}{c\rho} + \frac{1}{g}\ln \frac{1}{\tau} +1 +\frac{1}{g} \ln \frac{72}{\rho} + \frac{1}{g}\ln \frac{1}{\tau}}{(\frac{1}{3}-2c)\cdot \tau/3} \Bigg\}. \label{sdefecepsas7hepsilon}
\end{align}
Ignoring the constants in (\ref{sdefecepsas7hepsilon}), we have
\begin{align}
\frac{1}{h(\epsilon,g,\rho) }   & = O\bigg(\frac{1}{\tau}\ln \frac{1}{\tau}\bigg).\label{sdefecepsas7hepsilonhg}
\end{align}
As noted, $h(\epsilon,g,\rho)$ is a lower bound of ${\epsilon'}$, which implies that $1/[h(\epsilon,g,\rho)]$ is an upper bound of ${1}/{\epsilon'}$ (i.e., the term in (\ref{epstau3da})). Then from (\ref{sdefecepsas7hepsilonhg}),
the term in (\ref{epstau3da}) can be expressed as $O\big(\frac{1}{\tau}\ln \frac{1}{\tau}\big)$. From $\epsilon=\tau/3$, this means that $n$ specified in (\ref{Tgammaietajboundn2}) becomes $n = O\big(\frac{B \ln (1/\tau)}{\tau^2 }\cdot \ln \frac{m}{\beta}\big)$.
Hence, the proof of Theorem \ref{thm-entire-execution} is now completed.
\pfe

\section{Related Work}  \label{related}

In many practical applications, statistical learning is often adaptive---the queries on a dataset depend on previous interactions with the same dataset. However, generalization guarantees are  traditionally given in a non-adaptive model. Recent studies by Hardt and Ullman \cite{hardt2014preventing} as well as Dwork~\emph{et~al}.~\cite{dwork2015generalizationarXiv,dwork2015preserving,dwork2015Sciencereusable} provide generalization bounds in adaptive statistical learning, while Dwork \emph{et al}. \cite{dwork2015generalizationarXiv,dwork2015preserving,dwork2015Sciencereusable} also propose mechanisms via differential privacy and max-information. Differential privacy means that an adversary
given access to the output does not have much confidence to determine
whether the output was sampled from the probability distribution
generated by the algorithm under a database   $x$ or under a neighboring database~$x'$ that differs from $x$ in one record.  Specifically, a randomized algorithm $Y$ satisfies $\epsilon$-differential privacy
if for all neighboring databases $x$, $x'$ and any subset $\mathcal{Y}$ of the ouput range of the mechanism $Y$, it holds that
\mbox{$\bp{Y(x)\in \mathcal{Y}}\leq e^{\epsilon}\bp{Y(x')\in \mathcal{Y}}.$}  The notion of max-information gives generalization since it upper bounds the probability of ``bad events'' that can occur as a result of the dependence of the output variable $Y(X)$ on the input variable~$X$.

Rogers \emph{et al}. \cite{rogers2016max} show the connection between approximate differential privacy and
max-information,
and prove that the connection holds only for data
drawn from product distributions,
 where approximate differential privacy (ADP) \cite{dwork2006our} relaxes differential privacy so that $(\epsilon,\delta)$-ADP means the probabilities that the same output is seen on neighboring databases $x$ and $x'$ (differing in one record) is bounded by a factor $e^{\epsilon}$, in addition to a small additive probability $\delta$; i.e., $\bp{Y(x)\in \mathcal{Y}}\leq e^{\epsilon}\bp{Y(x')\in \mathcal{Y}}+\delta$ for any subset $\mathcal{Y}$ of the ouput range of the mechanism $Y$. Very recently, Bassily and Freund \cite{bassily2016typicality} propose an algorithmic stability notion called \emph{typical stability} which provides generalization  for a
broader class of queries than that of bounded-sensitivity queries (bounded sensitivity is often required by differential privacy). Typical stability means that the output of a query is ``well-concentrated''
around its expectation with respect to the underlying distribution on the dataset. Cummings \emph{et al}. \cite{cummings2016adaptive}   introduce different generalization notions and discuss their relationships with differential privacy. Russo and Zou \cite{russo2016controlling} present a  mutual-information framework for adaptive statistical learning and compare it with max-information.
 Blum and Hardt \cite{blum2015ladder} design an algorithm to maintain an accurate leaderboard for machine learning competitions (such as those
organized by Kaggle Inc. at \url{http://www.kaggle.com/}\,), where submissions can be adaptive.

Since differential privacy was proposed to quantify privacy analysis \cite{Dwork2006,dwork2006calibrating}, this notion has received much attention in the literature \cite{blocki2016differentially,tramer2015differential,qin2016heavy,Jiang:2013:PTD:2484838.2484846,KamalikaChaudhuriAllerton17}. Kifer and Machanavajjhala \cite{kifer2011no} observe that differential privacy may not work well when the data tuples are correlated in between. To generalize differential privacy,
Kifer and Machanavajjhala \cite{kifer2012rigorous} introduce the Pufferfish framework by considering the generation of the database and the adversarial belief about the database. A subclass of
the Pufferfish framework, called the Blowfish framework, is investigated by He \emph{et al}. \cite{he2014blowfish}. Blowfish privacy imposes deterministic
policy constraints rather than probabilistic correlations to model adversarial
knowledge. A general mechanism to achieve Pufferfish privacy is recently proposed by Song~\emph{et~al}.~\cite{wang2016privacy}. Xiao and Xiong~\cite{xiao2015protecting} incorporate temporal correlations into differential privacy    in the context of location privacy.
Chen \emph{et al}. \cite{chen2014correlated} and Zhu~\emph{et~al}.~\cite{zhu2015correlated} give different algorithms for privacy under data correlations. To improve the utilities of these algorithms, Liu~\emph{et~al}.~\cite{Changchang2016} present a Laplace mechanism that
 handles pairwise correlations. Yang~\emph{et~al}.~\cite{yang2015bayesian} consider different   adversary models and formalize the notion of Bayesian differential privacy that tackles tuple correlations as well. Yang \emph{et al}. \cite{yang2015bayesian} further introduce a mechanism that is only for the sum query on a Gaussian Markov random field with positive correlations and its extension to a discrete domain. For Bayesian differential privacy, Zhao \emph{et al}. \cite{BDPIT}   present mechanisms for databases with arbitrary tuple correlations and elaborate the case of tuple correlations being modeled by a Markov chain.

\section{Conclusion}
\label{sec:Conclusion}
 Recently, Dwork \emph{et al}. \cite{dwork2015generalizationarXiv,dwork2015preserving,dwork2015Sciencereusable} show that the holdout dataset from i.i.d. data samples can be reused in adaptive statistical learning, if the estimates are
perturbed and coordinated using techniques developed for differential privacy, which is a widely used notion to define privacy. Yet, the results of Dwork~\emph{et al}. \cite{dwork2015generalizationarXiv,dwork2015preserving,dwork2015Sciencereusable} are applicable to only the case of i.i.d. samples. In this paper, we show that Bayesian differential privacy can be used to ensure statistical
validity in adaptive statistical learning, where Bayesian differential privacy is introduced by Yang \textit{et al.}  \cite{yang2015bayesian} to extend differential privacy for addressing the case when data records are correlated. Specifically, we prove that the holdout dataset from \emph{correlated samples} can be reused in adaptive statistical learning, if the estimates are
perturbed and coordinated using techniques satisfying Bayesian differential privacy. Our results generalize those of Dwork \emph{et al}. \cite{dwork2015generalizationarXiv,dwork2015preserving,dwork2015Sciencereusable}  for i.i.d. samples to arbitrarily correlated data.

\linespread{1}

 { \fontsize{9}{10} \selectfont

\let\OLDthebibliography\thebibliography
 \renewcommand\thebibliography[1]{
   \OLDthebibliography{#1}
   \setlength{\parskip}{.5pt}
 }




}

$~$

\noindent \textbf{\Large Appendix} \vspace{-10pt}

\setcounter{section}{1}

\renewcommand\thesection{\Alph{section}}
\renewcommand\thesubsection{\thesection.\arabic{subsection}}

\subsection{Establishing Theorem~\ref{thm-BDP-max-information}} \label{secprfthm-BDP-max-information}

To begin with, we first define \textit{martingale} and \textit{Doob martingale} that will be used in the proof of Theorem~\ref{thm-BDP-max-information}. A sequence of
random variables $Z_0,Z_1,\ldots$ is referred to as a \textit{martingale} \cite{colantonio2009probabilistic} if $\mathbb{E}[Z_{i+1} | Z_0,\ldots,Z_i] = Z_i$ and $\mathbb{E}[|Z_i|] < \infty $ for $i = 0, 1, \ldots$, where $\mathbb{E}[\cdot]$ stands for the expected value of a random variable. A \textit{Doob martingale} \cite{colantonio2009probabilistic} is a martingale constructed
using the following general approach. Let $X_0,X_1,\ldots,X_m$ be a sequence of random variables,
and let $Y$ be a random variable with $\mathbb{E}[|Y|] <\infty $ (In general, $Y$ is a function of
$X_0,X_1,\ldots,X_m$). Then
the sequence consisting of $Z_i = \mathbb{E}[Y | X_0,\ldots,X_i]   $ for $i=0,1,\ldots,m$
gives a Doob martingale. Note that in the expression $Z_i = \mathbb{E}[Y | X_0,\ldots,X_i]   $, the expectation is only taken over $X_{i+1},X_{i+2},\ldots,X_m$ while  $X_0,\ldots,X_i$ are kept as random variables, so $Z_i$ is still a random variable.

The Azuma-Hoeffding inequality (see Lemma 2 of \cite{bollobas2001degree}) presented below is widely used in the analysis of   martingales.

\begin{lem}[Azuma-Hoeffding Inequality as Lemma 2 of \cite{bollobas2001degree}] \label{AHineq}
If $Z_i|_{i=0,\ldots,n}$ is a   martingale such that $|Z_i-Z_{i-1}|
\leq c$ for each $i=1,\ldots,n$, then $\bp{Z_n-Z_0\geq t} \leq \exp\big(\frac{-t^2}{2nc^2}\big)$.
\end{lem}

We now continue the proof ofTheorem~\ref{thm-BDP-max-information}.
We first fix $y \in \mathcal{Y}$, and define a function $g(x) \de \ln \left( \frac{\mathbb{P}[Y=y \boldsymbol{\mid} X=x]}{\mathbb{P}[Y=y ]}\right)$.
For function $g$, we will define a Doob martingale $Z_i|_{i=0,\ldots,n}$ with respect to $X_i|_{i=1,\ldots,n}$. Specifically, we define $$Z_i(x_1,\ldots,x_i)\de \mathbb{E}_{X_{i+1},\ldots,X_n}[g(x_1,\ldots,x_i,X_{i+1},\ldots,X_n)]$$ for $i=0,\ldots,n$; i.e., given $x_1,\ldots,x_i$, we take the expectation of \\$g(x_1,\ldots,x_i,X_{i+1},\ldots,X_n)$ with respect to $X_{i+1},\ldots,X_n$, and obtain\\ $Z_i(x_1,\ldots,x_i)$. For simplicity, we write the sequence $X_{i+1},\ldots,X_n$ as $X_{i+1:n}$, and write the sequence $x_1,\ldots,x_i$ as $x_{1:i}$, and so on. Then
\begin{align}
Z_i(x_{1:i}) & = \mathbb{E}_{X_{i+1:n}}[g(x_{1:i},X_{i+1:n})]   \label{eqn-Zi-X1i0} \\ & = \sum_{x_{i+1:n}} g(x_{1:i},x_{i+1:n})\bp{x_{i+1:n} \con x_{1:i}}. \label{eqn-Zi-X1i}
\end{align}
Replacing $i$ with $i-1$ in (\ref{eqn-Zi-X1i}), we obtain
\begin{align}
 & Z_{i-1}(x_{1:i-1})\nonumber \\ &  = \mathbb{E}_{X_{i:n}}[g(x_{1:{i-1}},X_{i:n})]  \nonumber \\ & = \sum_{x_{i}'} \sum_{x_{i+1:n}} g(x_{1:i-1},x_{i}',x_{i+1:n})\bp{x_{i}',x_{i+1:n}\con
x_{1:i-1}}. \label{eqn-Zi-X1iminus1}
\end{align}
To find the connection between (\ref{eqn-Zi-X1i}) and (\ref{eqn-Zi-X1iminus1}), we first note \\$\bp{x_{i}',x_{i+1:n}\con
x_{1:i-1}}= \bp{x_{i+1:n} \con x_{1:i-1},x_{i}'} \cdot  \bp{x_{i}'\con x_{1:i-1}}$ by the chain rule; put it in detail, we have
\begin{align}
 & \bp{x_{i}',x_{i+1:n}\con
x_{1:i-1}} \nonumber \\ &
= \frac{\bp{x_{1:i-1},x_{i}',x_{i+1:n}}}{\bp{x_{1:i-1}}}  \nonumber \\ &
= \frac{\bp{x_{1:i-1},x_{i}',x_{i+1:n}}}{\bp{x_{1:i-1},x_{i}'}}
\cdot \frac{\bp{x_{1:i-1},x_{i}'}}{\bp{x_{1:i-1}}}   \nonumber \\ &
= \bp{x_{i+1:n} \con x_{1:i-1},x_{i}'} \cdot  \bp{x_{i}'\con x_{1:i-1}}. \label{eqn-Zi-X1iminus1b}
\end{align}
Substituting (\ref{eqn-Zi-X1iminus1b})  into (\ref{eqn-Zi-X1iminus1}), we establish
\begin{align}
 & Z_{i-1}(x_{1:i-1})  \nonumber \\ &= \mathbb{E}_{X_{i:n}}[g(x_{1:{i-1}},X_{i:n})]  \nonumber \\ & = \sum_{x_{i}'} \sum_{x_{i+1:n}} \begin{array}{l} g(x_{1:i-1},x_{i}',x_{i+1:n})\bp{x_{i+1:n} \con x_{1:i-1},x_{i}'} \\ \cdot  \bp{x_{i}'\con x_{1:i-1}} \end{array} \nonumber \\ &= \sum_{x_{i}'} \big\{\mathbb{E}_{X_{i+1:n}}[g(x_{1:i-1},x_{i}',X_{i+1:n})] \cdot  \bp{x_{i}'\con x_{1:i-1}}\big\}. \label{eqn-Zi-X1iminus1c}
\end{align}
In view of (\ref{eqn-Zi-X1i0}) and
(\ref{eqn-Zi-X1iminus1c}), we will prove $|Z_i-Z_{i-1}|
\leq \epsilon$ by showing
for any $x_{i}'$ that
\begin{align}
|\mathbb{E}_{X_{i+1:n}}[g(x_{1:i-1},x_{i}',X_{i+1:n})]-\mathbb{E}_{X_{i+1:n}}[g(x_{1:i},X_{i+1:n})] |\leq \epsilon. \label{eqn-Zi-X1iminus1d}
\end{align}
We have
\begin{align}
&\mathbb{E}_{X_{i+1:n}}[g(x_{1:i-1},x_{i}',X_{i+1:n})] = \ln \frac{\mathbb{P}[Y(x_{1:i-1},x_{i}',X_{i+1:n})=y ]}{\mathbb{P}[Y=y ]} \label{eqn-Zi-X1iminus1d1}
\end{align}
and
\begin{align}
&\mathbb{E}_{X_{i+1:n}}[g(x_{1:i},X_{i+1:n})] = \ln \frac{\mathbb{P}[Y(x_{1:i},X_{i+1:n})=y ]}{\mathbb{P}[Y=y ]} \label{eqn-Zi-X1iminus1d2}.
\end{align}
Under $\epsilon$-Bayesian differential privacy, it holds by definition that $e^{-\epsilon} \leq  \frac{\mathbb{P}[Y(x_{1:i-1},x_{i},X_{i+1:n})=y ]}{\mathbb{P}[Y(x_{1:i-1},x_{i}',X_{i+1:n})=y ]} \leq e^{\epsilon}$, which   with (\ref{eqn-Zi-X1iminus1d1}) and  (\ref{eqn-Zi-X1iminus1d2}) implies  (\ref{eqn-Zi-X1iminus1d}). Since (\ref{eqn-Zi-X1iminus1d}) holds for any $x_{i}'$, we then obtain $|Z_i-Z_{i-1}|
\leq \epsilon$ from (\ref{eqn-Zi-X1i0}) and
(\ref{eqn-Zi-X1iminus1c}). Since $i$ can iterate through $\{1,\ldots,n\}$, we have proved $|Z_i-Z_{i-1}|
\leq \epsilon$ for $i=1,\ldots,n$. Then we use the Azuma-Hoeffding inequality (i.e., Lemma \ref{AHineq} above) and obtain
$\bp{Z_n-Z_0\geq t} \leq \exp\big(\frac{-t^2}{2n{\epsilon}^2}\big)$ for any $t>0$. By definition, $Z_n(x_{1:n})  = g(x_{1:n}) = g(x) $ and $Z_0  = \mathbb{E}_{X_{1:n}}[g(X_{1:n})] = \mathbb{E}_{X}[g(X)] $. Hence, it follows that
\begin{align}
\bp{g(x)-\mathbb{E}_{X}[g(X)]\geq t}  &\leq \exp\bigg(\frac{-t^2}{2n{\epsilon}^2}\bigg) . \label{ZnZ0bound}
\end{align}

We now evaluate $\mathbb{E}_{X}[g(X)]$ as follows.
\begin{align}
\mathbb{E}_{X}[g(X)] & =\mathbb{E}_{X}\bigg[\ln\frac{\bp{Y(X)=y}}{Y=y}\bigg]\nonumber \\ &=\mathbb{E}_{X}[\ln \bp{Y(X)=y}]-\mathbb{E}_{X}[\ln \bp{Y=y}]  \nonumber \\ & = \mathbb{E}_{X}[\ln \bp{Y(X)=y}]-\ln \bp{Y=y}.\label{eqexgx1}
\end{align}
 Since the natural logarithm   $\ln $ is a convex function, we use Jensen's inequality to obtain
\begin{align}
 & \mathbb{E}_{X}[\ln \bp{Y(X)=y}]  \leq \ln [\mathbb{E}_{X}\bp{Y(X)=y}]=\ln \bp{Y=y}. \label{eqexgx2}
\end{align}
Using (\ref{eqexgx2}) in  (\ref{eqexgx1}), we have $\mathbb{E}_{X}[g(X)]\leq 0$, which along with (\ref{ZnZ0bound}) further yields
\begin{align}
\bp{g(x) \geq t}  &\leq \exp\bigg(\frac{-t^2}{2n{\epsilon}^2}\bigg) . \label{ZnZ0bound2}
\end{align}

For an integer $i \geq 1$, we define $t_i \de 2{\epsilon}^2 n +  \epsilon \sqrt{2 n \ln (2^i/\beta)}$ and define $B_i \de \big\{x \con t_i <g(x)\leq t_{i+1}  \big\}$. Let $B_y \de  \big\{x \con g(x) > t_1 \big\} = \bigcup_{i \geq 1}B_i$.
By Bayes' rule, for every $x \in B_i$, it holds that
\begin{align}
\frac{\mathbb{P}[X=x \boldsymbol{\mid} Y=y]}{\mathbb{P}[X=x]}
 &= \frac{\mathbb{P}[Y=y \boldsymbol{\mid} X=x]}{\mathbb{P}[Y=y]}  = \exp\big(g(x)\big)
\leq \exp(t_{i+1}). \label{YxBaeys}
\end{align}
Therefore, we obtain from (\ref{ZnZ0bound2}) and (\ref{YxBaeys}) that
\begin{align}
&\mathbb{P}[X \in B_i \boldsymbol{\mid} Y=y]\nonumber \\
&= \sum_{x \in B_i}\mathbb{P}[X=x \boldsymbol{\mid} Y=y] \nonumber \\ & \leq \exp(t_{i+1}) \cdot \sum_{x \in B_i}\mathbb{P}[X =x]\nonumber \\
 & \leq  \exp(t_{i+1}) \cdot \mathbb{P}\big[g(X)\geq t_i\big]\nonumber \\ & \leq
\exp\left(t_{i+1}-\frac{{t_i}^2}{2 n {\epsilon}^2}\right) \nonumber \\
 &
= \exp\left(2\epsilon^2 n \hspace{-2pt}+\hspace{-2pt}\epsilon \sqrt{2 n \ln(2^{i+1}/\beta)}\hspace{-2pt}-\hspace{-2pt}\left(\hspace{-2pt}2\epsilon\sqrt{n}\hspace{-2pt}+\hspace{-2pt}\sqrt{2\ln(2^i/\beta)}\right)^2\hspace{-2pt}/2\hspace{-2pt}\right) \nonumber \\ & \leq  \exp\left(\epsilon\sqrt{2n}\left(
\sqrt{\ln(2^{i+1}/\beta)}-2\sqrt{\ln(2^{i}/\beta)}\right)-\ln(2^{i}/\beta)\right)\nonumber \\ & \leq  \exp\big(-\ln(2^i/\beta)\big)\nonumber \\ &
= \beta/2^i. \label{eqbta2i}
\end{align}
An immediate implication of (\ref{eqbta2i}) is that
\begin{align}
\mathbb{P}[X \in B_y \boldsymbol{\mid} Y=y]
&= \sum_{i\geq 1}\mathbb{P}[X \in B_i \boldsymbol{\mid} Y=y] \leq \sum_{i\geq 1} \beta/2^i \leq \beta.
\end{align}
Let $\mathcal{B}=\big\{(x,y)\con y \in \mathcal{Y},x \in B_y\big\}$. Then
\begin{align}
\mathbb{P}[(X,Y)\in \mathcal{B}]
&= \sum_{y \in \mathcal{Y}} \left\{\mathbb{P}[Y=y]\mathbb{P}[X \in B_y \boldsymbol{\mid} Y=y]\right\} \leq \beta.\label{eqbta2ia}
\end{align}
For every $(x, y) \not \in \mathcal{B}$, we have
\begin{align}
\mathbb{P}[X=x,Y=y]
&= \mathbb{P}[X=x\con Y=y] \cdot \mathbb{P}[Y=y] \nonumber \\ & \leq \exp(t_1) \cdot \mathbb{P}[X=x ] \cdot \mathbb{P}[Y=y],
\end{align}
and hence by (\ref{eqbta2ia}) we get
\begin{align}
\mathbb{P}_{(x, y) \sim p(X,Y)} \left[\frac{\mathbb{P}[X=x,Y=y]}{\mathbb{P}[X=x ] \cdot \mathbb{P}[Y=y]}\geq \exp(t_1)\right] \leq \beta.
\end{align}
This, by Lemma \ref{lemdwork2015generalizationarXiv} below (i.e., \cite[Lemma 18]{dwork2015generalizationarXiv}), gives that
\begin{align}
 &I_{\infty}^{\beta}(X;Y)   \leq \log\big(\exp(t_1)\big) = t_1 \log e =\big(2\epsilon^2n +\epsilon\sqrt{2n \ln(2/\beta) }\,\big)\log e,\nonumber
\end{align}
  where $\log $ means the binary logarithm.
\pfe

\begin{lem} [\hspace{-.2pt}{\cite[Lemma 18]{dwork2015generalizationarXiv}\hspace{0pt}}\hspace{0pt}] \label{lemdwork2015generalizationarXiv}
Let $X$ and $Y$ be two random variables over the same domain $\mathcal{X}$. If $\mathbb{P}_{x \sim p(X)} \left[\frac{\mathbb{P}[X=x]}{ \mathbb{P}[Y=x]}\geq 2^k\right] \leq \beta$, then $I_{\infty}^{\beta}(X;Y)\leq k$.
\end{lem}

\def\x{\ref{cor-BDP-max-information}}
\def\y{\pageref{cor-BDP-max-information}}

\subsection{Proof of Lemma~\x~on Page \y} \label{secapcor-BDP-max-information}

We let $X \times Y$ be the random variable obtained by drawing
$X$ and $Y$ \emph{independently} from their probability distributions. We also define $\mathcal{O}$ as  the event that $|Y(X)-Y(\mathcal{P})|\geq \tau$. First, we use the concentration inequality to obtain $\bp{X \times Y\in \mathcal{O}}\leq 2e^{-{\tau^2}/({2n\cdot{\Delta_Q}^2})}$. This along with Theorem \ref{thm-BDP-max-information}  and the definition of $\beta$-approximate max-information $I_{\infty}^{\beta}(X;Y(X))$ implies that
for $k \de I_{\infty}^{\beta}(X;Y(X)) \leq \big(2\epsilon^2n +\epsilon\sqrt{2n \ln(2/\beta) }\,\big)\log e$, we bound the probability $\bp{(X,Y)\in \mathcal{O}} $ (i.e., $\bp{|Y(X)-Y(\mathcal{P})|\geq \tau}$) as follows:
\begin{align}
& \bp{|Y(X)-Y(\mathcal{P})|\geq \tau}\nonumber \\ & \leq 2^k \cdot 2e^{-{\tau^2}/({2n\cdot{\Delta_Q}^2})} + \beta \nonumber \\ & \leq 2^{ (2\epsilon^2n +\epsilon\sqrt{2n \ln(2/\beta) }\,)\log e} \cdot 2e^{-{\tau^2}/({2n\cdot{\Delta_Q}^2})}
+ \beta. \label{eqnYxYPbeta}
\end{align}

Suppose $\epsilon\leq a{\tau}/({n\Delta_Q})$ and $\beta=2e^{-{b\tau^2}/({ n\cdot{\Delta_Q}^2})}$ for some constants $a$ and $b$ that will be specified later. Then we obtain from (\ref{eqnYxYPbeta}) that
\begin{align}
& \bp{|Y(X)-Y(\mathcal{P})|\geq \tau}\nonumber \\ & \leq   2e^{(2a^2+a\sqrt{2b}-\frac{1}{2}){\tau^2}/({n\cdot{\Delta_Q}^2})}+ 2e^{-{b\tau^2}/({ n\cdot{\Delta_Q}^2})}. \label{eqnYxYPbeta2}
\end{align}
To ensure $2a^2+a\sqrt{2b}-{1}/{2}<0$ which implies $2a^2<{1}/{2}$ so that $a<{1}/{2}$, we set $a={1}/{3}$ for simplicity. Then $2a^2+a\sqrt{2b}-{1}/{2}=\frac{1}{3}\sqrt{2b}-\frac{5}{18}$. Solving $b$ for the objective function $\min_{b>0}\max\{\frac{1}{3}\sqrt{2b}-\frac{5}{18},-b\}$, we obtain $b\approx 0.12$. Hence, again for simplicity, we set $b=1/8$. Then $ 2a^2+a\sqrt{2b}-{1}/{2}=-1/9$. Applying the above to (\ref{eqnYxYPbeta2}), we have
\begin{align}
& \bp{|Y(X)-Y(\mathcal{P})|\geq \tau}\nonumber \\ & \leq   2 e^{-{\tau^2}/({9n\cdot{\Delta_Q}^2})}+ 2e^{-{\tau^2}/({8n\cdot{\Delta_Q}^2})}\nonumber \\ & \leq 4e^{-{\tau^2}/({9n\cdot{\Delta_Q}^2})}.\nonumber
\end{align}
~\vspace{-15pt}\pfe

\normalsize

\end{document}